\documentclass[sigconf]{acmart}
\AtBeginDocument{%
  }

\setcopyright{acmlicensed}
\copyrightyear{2018}
\acmYear{2018}
\acmDOI{XXXXXXX.XXXXXXX}
\acmConference[Conference acronym 'XX]{Make sure to enter the correct
  conference title from your rights confirmation email}{June 03--05,
  2018}{Woodstock, NY}
\acmISBN{978-1-4503-XXXX-X/2018/06}

\acmSubmissionID{1837}

\usepackage[table]{xcolor}
\usepackage{pdflscape}
\usepackage{soul}
\usepackage{makecell}
\usepackage{graphicx}
\usepackage{booktabs}
\usepackage{multirow}
\usepackage{rotating}
\usepackage{array}

\definecolor{zeroshotgray}{gray}{0.92}
\newcolumntype{Z}{>{\columncolor{zeroshotgray}}c}

\begin{document}

\title[LightCRS]{LightCRS: Lightweight Multi-Domain Conversational Recommendation System}

\author{He Ma}
\affiliation{%
  \institution{University of Sydney}
  \city{Sydney}
  \state{NSW}
  \country{Australia}}

\author{Lan You}
\affiliation{%
  \institution{University of Sydney}
  \city{Sydney}
  \state{NSW}
  \country{Australia}}

\author{Qiyang Wu}
\affiliation{%
  \institution{University of Sydney}
  \city{Sydney}
  \state{NSW}
  \country{Australia}}

\author{Chen Liu}
\authornote{Corresponding author.}
\email{liu.chen@nankai.edu.cn}
\affiliation{%
  \institution{Nankai University}
  \city{Tianjin}
  \country{China}}

\renewcommand{\shortauthors}{Ma et al.}

\begin{abstract}
  Conversational recommender systems have recently adopted vision-language models to match user dialogue intents with multimodal product information. However, extending these methods to many product domains faces a dilemma: deploying a separately adapted VLM per domain scales cost linearly with the number of domains and requires routing each dialogue to the right model, while naively merging all domains into a single VLM degrades accuracy because product characteristics vary widely across categories. We propose LightCRS, a lightweight, query-conditioned framework for unified multi-domain multimodal conversational recommendation. LightCRS precomputes multimodal product representations once offline with frozen encoders. Online, a compact 5.3M-parameter module fuses this evidence for each dialogue via query-conditioned cross-attention, whose parameters are shared across all domains, and compares candidates jointly with an inter-candidate Transformer. To enable rigorous evaluation, we construct MM-R2A-25, a 25-domain benchmark pairing natural Reddit recommendation-seeking conversations with Amazon product listings. LightCRS achieves state-of-the-art average HR@1, NDCG@5, and MRR, improving HR@1 by 35.4\% over the strongest baseline while reducing per-query latency by 266×. Joint multi-domain training itself improves HR@1 by 15.2\% over separately trained single-domain models.
\end{abstract}






\maketitle

\section{Introduction}
Conversational recommender systems (CRSs) predict items that satisfy user intents expressed through natural-language dialogue~\cite{xie2024neighborhood,he2024reindex,yue2025preference,wei2025mscrs,jeon2025adapting,zhang2026moon}. Recent progress in CRSs has been driven by vision-language models (VLMs), which integrate visual and textual product evidence to improve product understanding~\cite{jeon2025adapting,zhang2026moon,giahi2025vl}. Meanwhile, e-commerce platforms host a large and growing number of product categories, and maintaining a separately adapted VLM for each category is untenable in practice. Therefore, a unified multimodal CRS that serves multiple domains is desirable.


Existing multimodal CRSs, however, face a dilemma when extended to many domains. Deploying per-domain models~\cite{yue2025preference,wei2025mscrs,jeon2025adapting} scales training and inference costs approximately linearly with the number of domains, and introduces a routing problem: the system must decide which model should handle each dialogue, which is difficult when the relevant domain is ambiguous. Conversely, naively merging all domains into a single VLM can degrade recommendation accuracy, because product attributes, visual characteristics, and recommendation criteria vary widely across categories~\cite{jeon2025adapting}. Nor can task-specific training be skipped: general-purpose proprietary models and VLMs applied through zero-shot prompting perform poorly on conversational recommendation~\cite{jeon2025adapting, xi2024memocrs, huang2025recommender, wang2024recmind, liao2026eliminating}.


To address these limitations, we propose LightCRS, a lightweight, query-conditioned, multi-domain framework that resolves this dilemma. Following the candidate-based CRS setting of LaViC~\cite{jeon2025adapting}, LightCRS ranks a set of candidates for each dialogue through two stages. Offline, since candidate information is static, each product is represented by five views: title, features, description, image captions, and visual tokens. These representations are precomputed once with frozen encoders, which removes VLM inference from the online path.
Online, a compact 5.3M-parameter trainable module scores candidates with two components: a query-conditioned cross-attention that fuses the five views for each dialogue across 25 domains, and an inter-candidate Transformer that lets candidates be compared jointly rather than scored in isolation.

The key observation behind our design is that which product evidence matters is determined by the dialogue itself: a user describing a product's appearance needs visual signals, whereas one asking about CPU specifications needs textual attributes. Query-conditioned cross-attention therefore allows a single set of shared parameters to behave differently for every dialogue. This addresses all three obstacles at once. A single unified model eliminates per-domain replication and the routing mechanism it requires; offline precomputation with a compact online module keeps training and inference costs low; and sharing the attention parameters across categories turns joint multi-domain training from a liability into an asset, as our multi-domain model outperforms 25 separately trained single-domain counterparts by 15.2\% in HR@1. 
Overall, LightCRS improves average HR@1 by 35.4\% over the strongest baseline while reducing per-query latency by 266×. Although designed for supervised multi-domain recommendation, LightCRS also provides modest generalization gains on unseen product categories.

To enable rigorous evaluation, we introduce \textit{MM-R2A-25}, a 25-domain dataset linking conversational queries from Reddit recommendation communities to Amazon product listings. To the best of our knowledge, this is the first CRS benchmark designed to probe multi-domain visual and textual product understanding at scale.

Our main contributions are as follows:
\begin{itemize}
    \item We propose LightCRS, a lightweight multi-domain multimodal CRS that decouples offline product representation from online scoring: five textual and visual product views are precomputed with frozen encoders, and a compact trainable module combines query-conditioned cross-attention, which adaptively fuses five product views for each query while sharing parameters across all 25 domains, with an inter-candidate Transformer that captures relative relationships among candidates.

    \item We release MM-R2A-25, a 25-domain multimodal CRS benchmark that links natural Reddit recommendation dialogues to Amazon product listings with verified images.

    \item Extensive experiments show that LightCRS achieves state-of-the-art average HR@1, NDCG@5, and MRR, improving HR@1 by 35.4\% over the strongest baseline while reducing training and inference costs. Moreover, LightCRS benefits from joint multi-domain training, achieving a 15.2\% HR@1 improvement over single-domain training, and exhibits modest generalization to unseen categories.
\end{itemize}

\section{MM-R2A-25 Dataset}
\label{sec:dataset}

To evaluate multi-domain multimodal conversational recommendation at scale, we introduce \textit{MM-R2A-25}, a large-scale benchmark that follows the Reddit-Amazon construction protocol~\cite{jeon2025adapting} while extending its domain coverage to 25 diverse product categories. Unlike existing CRSs datasets, which are often limited in scale, domain coverage, or multimodal richness, \textit{MM-R2A-25} provides aligned Reddit conversations, comprehensive Amazon product metadata, and verified product images spanning all 25 categories.

Each conversation is drawn from Reddit communities\footnote{\url{https://pushshift.io/}} where users seek product recommendations,
and is paired with candidate product listings from the corresponding Amazon category\footnote{\url{https://amazon-reviews-2023.github.io/}}. Ground-truth items are identified through explicit product mentions or links within the final utterance of each retained thread.
Product images are verified and linked to the corresponding Amazon listings, enabling multimodal modeling with real visual signals. To construct this corpus, we apply GPT-3.5-turbo to each Reddit thread using the filtering template shown in Table~\ref{tab:prompt-filter} in Appendix~\ref{sec:appendix-prompts}. We retain only conversations labeled as ``yes'' by the model, corresponding to threads whose final utterance contains an explicit product recommendation; these retained instances form \textit{MM-R2A-25}. We further use LLaVA-v1.6 with the captioning prompt in Table~\ref{tab:prompt-caption} in Appendix~\ref{sec:appendix-prompts} to generate visually grounded product descriptions. 
The dataset spans a wide range of domain sizes ensuring that models are evaluated across both resource-rich and resource-sparse categories. The full per-category breakdown is provided in Table~\ref{tab:dataset} in Appendix~\ref{sec:appendix-dataset}. Table~\ref{tab:dataset-comparison} in Appendix~\ref{sec:appendix-dataset} further positions \textit{MM-R2A-25} relative to existing conversational recommendation benchmarks.

\section{Method}
\label{sec:method}

\begin{figure*}[t]
  \centering
  \includegraphics[width=\textwidth]{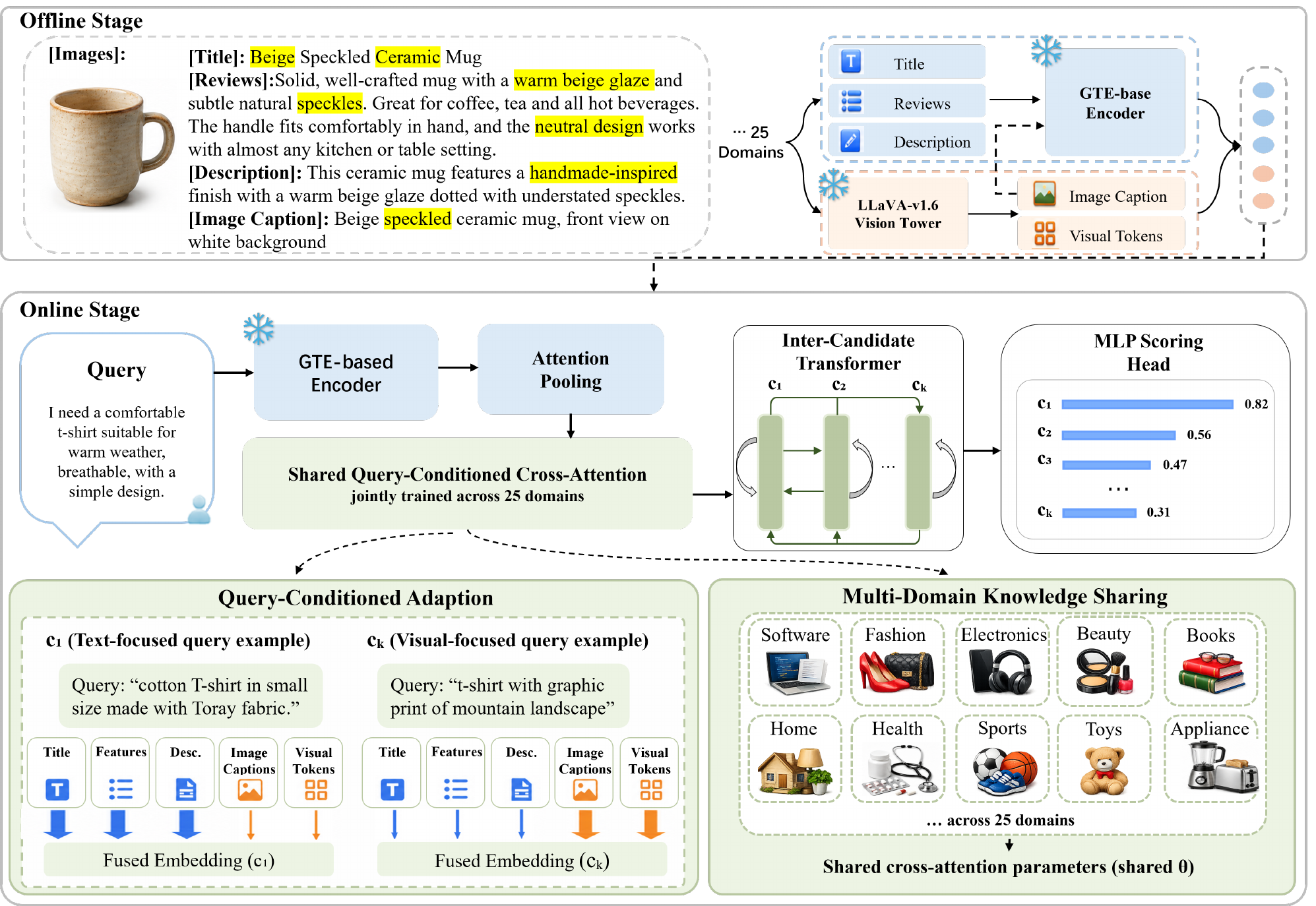}
  \caption{Architecture of LightCRS. The offline stage extracts and caches multi-view product representations using frozen GTE-base and LLaVA-v1.6. The online stage encodes the query, performs query-conditioned cross-attention over candidate representations, models candidate interactions with an inter-candidate Transformer, and outputs relevance scores through an MLP scoring head.}
  \label{fig:framework}
\end{figure*}

\subsection{Problem Statement}


We formulate multi-domain multimodal conversational recommendation as a re-ranking problem.
Following LaViC~\cite{jeon2025adapting}, we define the dialogue query at recommender turn $k$ as the prefix $q = C_{<k} = \{u_t\}_{t=1}^{k-1}$ containing every utterance before the recommender's $k$-th turn. The recommender's $k$-th utterance $u_k$, which mentions the recommended product, is treated as the ground-truth label only and is excluded from $q$, ensuring that the target product does not appear in the model input. Given $q$, an SBERT retriever~\cite{reimers2019sentence, he2023large} returns a candidate set $\mathcal{C} = \{c_1, c_2, \ldots, c_K\}$ of $K = 10$ candidates following the Reddit-Amazon setting~\cite{jeon2025adapting}, where each candidate $c_i$ denotes a product from one of multiple e-commerce domains. Each candidate set contains exactly one ground-truth item $c^+$ and $K-1$ negative candidates. Conversations associated with multiple ground-truth items are split into multiple instances. In addition, if $c^+$ is initially absent in the candidates, we swap out one negative candidate for $c^+$ to ensure precisely one ground-truth item. After candidate-set construction, we randomly shuffle all $K$ candidate items to remove the original retrieval order while prevent positional bias.
Each candidate is associated with heterogeneous multimodal product evidence, including its title $\mathbf{t}_i \in \mathbb{R}^{d_{\mathrm{txt}}}$, features $\mathbf{f}_i \in \mathbb{R}^{d_{\mathrm{txt}}}$, description $\mathbf{d}_i \in \mathbb{R}^{d_{\mathrm{txt}}}$, image captions $\mathbf{I}_i \in \mathbb{R}^{N \times d_{\mathrm{txt}}}$, and visual tokens $\mathbf{V}_i \in \mathbb{R}^{5 \times d_v}$, where $d_{\mathrm{txt}} = 768$ and $d_v = 4096$.
The goal is to learn a single multi-domain scoring model
$s(c_i \mid q, \mathcal{C}) \in \mathbb{R}$ that efficiently
integrates multimodal product evidence across diverse domains
and ranks candidates according to their alignment with the user's
conversational intent, assigning the ground-truth item
$c^+ \in \mathcal{C}$ a higher rank than all remaining candidates.
Furthermore, we focus on improving multi-domain recommendation accuracy, rather than generating fluent dialogue~\cite{jeon2025adapting, xie2024neighborhood, he2024reindex}.

\subsection{Proposed Framework}

In this section, we propose LightCRS, a lightweight, query-adaptive framework that aligns user dialogue with product representations across diverse domains. As illustrated in Figure~\ref{fig:framework}, LightCRS operates in two stages. 

\subsubsection{Offline Stage}
\label{sec:offline-features}
To improve computational efficiency, we precompute static candidates' multi-view representations offline. These cached representations are shared across training and evaluation. We next describe the encoding process for each view representation.

\paragraph{Text encoding.}
All three textual views, including the title, features, and description, are encoded by GTE-base~\cite{li2023towards}. Each view is mean-pooled over its token sequence into a single $d_{\mathrm{txt}}$-dimensional embedding:
title $\mathbf{t}_i \in \mathbb{R}^{d_{\mathrm{txt}}}$,
features $\mathbf{f}_i \in \mathbb{R}^{d_{\mathrm{txt}}}$,
description $\mathbf{d}_i \in \mathbb{R}^{d_{\mathrm{txt}}}$.

\paragraph{Image caption encoding.}
Product images are captioned offline using LLaVA-v1.6. The resulting captions are then encoded by the same GTE-base model, mean-pooled per caption, and stacked to form
$\mathbf{I}_i \in \mathbb{R}^{N \times d_{\mathrm{txt}}}$, where $N $ denotes the number of captions per product.

\paragraph{Visual-token extraction.}
Visual features are extracted with the vision tower of LLaVA-v1.6. For one representative image, we divide it into five spatial crops and encode each crop independently. The resulting crop-level visual features are projected into the language-model hidden space, yielding one $d_v$-dimensional visual token per crop. This yields five spatially grounded visual tokens, each summarizing a distinct region of the product image. The resulting visual representation is $\mathbf{V}_i \in \mathbb{R}^{5 \times d_v}$. 

The actual cost of the offline stage is reported in Appendix~\ref{sec:appendix-offline-cost}.

\subsubsection{Online Stage}
\label{sec:online-scoring}
At this stage, LightCRS uses a shared cross-attention module to align the on-the-fly encoded query with each candidate’s precomputed multi-view representations, enabling query-adaptive fusion and knowledge sharing across all 25 domains. An inter-candidate Transformer then compares candidates each other, and a scoring MLP outputs the final logits.

\paragraph{Query encoding.}
We tokenize the dialogue query with the same frozen GTE-base encoder into $T_q$ subword tokens, embed them as $\mathbf{Q} \in \mathbb{R}^{T_q \times d_{\mathrm{txt}}}$, project to $\mathbb{R}^{T_q \times D}$, and compress the resulting token sequence into a single vector $\mathbf{q} \in \mathbb{R}^{D}$ via learned attention pooling:
\begin{equation}
  \mathbf{q}
  = \operatorname{softmax}\Bigl(
      \frac{\mathbf{w}_q(\mathbf{Q}\mathbf{W}_p)^\top}{\sqrt{D}}
    \Bigr)
    (\mathbf{Q}\mathbf{W}_p),
\end{equation}
with $\mathbf{W}_p \in \mathbb{R}^{d_{\mathrm{txt}} \times D}$ and $\mathbf{w}_q \in \mathbb{R}^{1 \times D}$ learnable, yielding $\mathbf{q} \in \mathbb{R}^D$.

\paragraph{Query-conditioned cross-attention.}
For each candidate $c_i$, we project all five feature views into a shared $D$-dimensional space via learnable linear projections $\hat{\mathbf{t}}_i = \mathbf{W}_t \mathbf{t}_i$, $\hat{\mathbf{f}}_i = \mathbf{W}_f \mathbf{f}_i$, $\hat{\mathbf{d}}_i = \mathbf{W}_d \mathbf{d}_i$, $\hat{\mathbf{I}}_i = \mathbf{I}_i \mathbf{W}_I$, and $\hat{\mathbf{V}}_i = \mathbf{V}_i \mathbf{W}_V$, and concatenate them:
\begin{equation}
  \mathbf{C}_i = \bigl[\,\hat{\mathbf{t}}_i;\; \hat{\mathbf{f}}_i;\; \hat{\mathbf{d}}_i;\; \hat{\mathbf{I}}_i;\; \hat{\mathbf{V}}_i\,\bigr] \in \mathbb{R}^{T_c \times D},
\end{equation}
where $T_c = 3 + N + 5$. A binary attention mask zeroes out padded positions for image-caption, handling the variable number of images per product.

The query vector is replicated across all $K$ candidates and processed through $L$ cross-attention layers, where each layer applies multi-head cross-attention followed by a feed-forward network, both with post-norm residuals:
\begin{align}
  \mathbf{q}^{(l)}_i
  &= \mathrm{LN}\,\bigl(
      \mathbf{q}^{(l-1)}_i
      + \mathrm{MHA}\,\bigl(
          \mathbf{q}^{(l-1)}_i,
          \mathbf{C}_i,
          \mathbf{C}_i
        \bigr)
    \bigr), \\
  \mathbf{q}^{(l)}_i
  &= \mathrm{LN}\,\bigl(
      \mathbf{q}^{(l)}_i
      + \mathrm{FFN}\,\bigl(\mathbf{q}^{(l)}_i\bigr)
    \bigr).
\end{align}
Rather than committing a priori to a fixed weighting of modalities, the cross-attention splits its weights across all five views in proportion to how relevant each view is to the current dialogue. This lets the model adapt to each user's intent within a domain: it attends more to visual tokens when a user describes the appearance of a product, and more to textual descriptions when the user asks about CPU or GPU type of a computer. The same product can therefore be scored differently depending on the information highlighted in the dialogue.

Beyond query-specific view weighting, cross-attention also serves as the shared mechanism for multi-domain knowledge fusion. All 25 domains jointly update the same attention parameters, allowing domain-specific modality preferences to be consolidated into a unified query-view alignment function. 
For example, queries in text-oriented domains such as Software tend to emphasize titles, features, and descriptions, whereas queries in visually oriented domains such as Handmade place greater weight on image captions and visual tokens. Because all domains share the same cross-attention parameters, query–view alignment patterns learned from one domain can transfer to others, helping the model interpret query intent and identify the most relevant product views. This design enables a single cross-attention module to support all 25 domains.

\paragraph{Inter-candidate comparison.}
The fusion above scores each candidate independently against the query. To decide which candidate is the best, the per-candidate fused representations $\mathbf{q}_i^{(L)}$, which now carry the query-conditioned multi-view evidence of candidate $c_i$, are exposed to each other through a Transformer encoder. Stacking these across the $K$ candidates gives $\mathbf{F} = [\,\mathbf{q}_1^{(L)};\; \mathbf{q}_2^{(L)};\; \dots;\; \mathbf{q}_K^{(L)}\,] \in \mathbb{R}^{K \times D}$, which is fed to the encoder as
\begin{equation}
  \mathbf{H}
  = \mathrm{TransformerEncoder}\,\bigl(\mathbf{F}\bigr)
  \in \mathbb{R}^{K \times D},
\end{equation}
whose $i$-th row $\mathbf{h}_i \in \mathbb{R}^D$ is the comparison-aware representation of candidate $c_i$: it preserves the query-conditioned multi-view evidence from $\mathbf{q}_i^{(L)}$ and is refined through inter-candidate self-attention. 

\paragraph{Scoring head.}
The representation of each candidate is then mapped to a scalar logit by a two-layer MLP with GELU activation:
\begin{equation}
  s_i
  = \mathbf{w}_2^\top
    \mathrm{GELU}\,\bigl(
      \mathbf{W}_1 \mathbf{h}_i + \mathbf{b}_1
    \bigr)
    + b_2,
\end{equation}
with $\mathbf{W}_1 \in \mathbb{R}^{D/2 \times D}$, $\mathbf{b}_1 \in \mathbb{R}^{D/2}$, $\mathbf{w}_2 \in \mathbb{R}^{D/2}$, and $b_2 \in \mathbb{R}$ learnable. Stacking the $K$ candidates yields the logit vector $\mathbf{s} = [s_1, \ldots, s_K]^\top \in \mathbb{R}^K$, which is the model's output $s(\cdot \mid q, \mathcal{C})$ for the given query and candidate set.

\subsection{Training Objective}

We train the full model end-to-end with a cross-entropy loss over the $K$-way candidate set. The scoring head produces logits $\mathbf{s} \in \mathbb{R}^K$, and we treat reranking as $K$-way classification in which the target is the index $y^+$ of the ground-truth candidate $c^+$:
\begin{equation}
  \mathcal{L}_{\mathrm{ce}} = -\log \frac{\exp(s_{y^+})}{\sum_{j=1}^{K} \exp(s_j)}.
\end{equation}

\section{Evaluations}

\subsection{Experimental Setup}
\label{sec:experimental-setup}

\paragraph{Dataset.}
We evaluate all methods on the newly constructed \textit{MM-R2A-25} dataset, which spans 25 product domains. Section~\ref{sec:dataset} details the dataset construction, while Table~\ref{tab:dataset} summarizes its statistics. 
For each domain, we partition the conversations into disjoint training, validation, and test sets using an 8:1:1 ratio.

\paragraph{Baselines.}
We compare our proposed method with three baseline categories: task-specific CRS methods, generative methods, and retrieval methods. Although earlier knowledge graph-based CRSs methods have shown effectiveness~\cite{chen2019towards,zhou2020improving,wang2022towards}, they rely heavily on domain-specific knowledge graphs and have mainly been evaluated in movie recommendation. Generalizing such approaches consistently to our 25 diverse product domains is non-trivial. Moreover, prior studies have shown that zero-shot large language models often outperform knowledge graph-based conversational recommender systems on Reddit dialogues~\cite{he2023large}. Therefore, we focus on these three baseline categories because they are relevant to candidate-based conversational recommendation and applicable across multiple product domains.

(1) Task-specific CRS methods.
These methods are specifically designed or adapted for conversational recommendation. Among the multimodal methods, LaViC~\cite{jeon2025adapting} distills visual knowledge from LLaVA and finetunes the resulting model with LoRA. MSCRS~\cite{wei2025mscrs} integrates collaborative, textual, and visual semantic graphs through GNN-based prompt learning. PRIME~\cite{yue2025preference} adopts a retrieval--reranking pipeline in which a Qwen2-VL-2B multimodal ranker reranks candidates using online preference feedback. Among methods without direct visual signals, RTA~\cite{he2024reindex} reindexes item titles as individual tokens and calibrates item-level generation probabilities, while NBCRS~\cite{xie2024neighborhood} models collaborative preferences through item neighborhoods.

(2) Generative methods.
These methods take the conversation and candidate-item information as input and use shared prompt templates to select a single most relevant item from the candidate list. 
MiniMax-M2.7 and MiniMax-M3 are proprietary models accessed through APIs. Vicuna-v1.5~\cite{chiang2023vicuna} is a text-only instruction-following LLM. LLaVA-v1.5~\cite{liu2024improved} extends Vicuna through visual instruction tuning, while LLaVA-v1.6~\cite{liu2024llavanext} further improves multimodal reasoning and multi-image processing.

(3) Retrieval methods.
These methods rank candidate items according to the similarity between the conversation and item titles. BM25~\cite{robertson2009probabilistic} performs lexical matching based on token overlap. SBERT~\cite{reimers2019sentence} generates sentence embeddings using a siamese BERT architecture. RoBERTa\textsubscript{large}~\cite{liu2019roberta} uses representations from a large pretrained language encoder, while SimCSE\textsubscript{large}~\cite{gao2021simcse} learns sentence embeddings through contrastive training. 

\paragraph{Metrics.}
We use \textit{HitRatio@1 (HR@1)} , \textit{NormalizedDiscountedCumulativeGain@5 (NDCG@5)}, and \textit{MeanReciprocalRank (MRR)} as recommendation accuracy metrics. HR@1 directly measures how often the model identifies the correct item as the best candidate. NDCG@5 evaluates whether the ground-truth item is placed within a top-ranked shortlist. MRR further considers the exact position of the ground-truth item across the entire candidate list. Beyond recommendation accuracy, we report both training and inference costs to evaluate whether performance improvements are achieved. Trainable parameters measure the optimization burden during task adaptation, whereas total parameters and on-disk model size reflect the complete deployment footprint. For inference, peak GPU memory and per-query MFLOPs characterize memory and computational demand, while wall-clock latency (ms/query) and throughput (queries/second) measure end-to-end serving efficiency. 

\paragraph{Implementation Details.}
All locally executed training and inference experiments are conducted on a single NVIDIA DGX Spark workstation equipped with an NVIDIA GB10 Grace Blackwell Superchip and 128~GB of unified system memory. Using a single workstation provides a consistent hardware environment for comparing the computational costs of LightCRS and baselines. In addition, we obtain open-source pretrained models (SBERT, RoBERTa\textsubscript{large}, SimCSE\textsubscript{large}, Vicuna, and LLaVA) from the official HuggingFace\footnote{https://huggingface.co/} repositories, while MiniMax-M3 and MiniMax-M2.7 are accessed through their official APIs and therefore are excluded from hardware-dependent efficiency comparisons. Moreover, for all methods, we report the mean and standard deviation of HR@1, NDCG@5, and MRR over three independent runs. 
Retrieval methods (BM25, SBERT, RoBERTa, and SimCSE) are deterministic given fixed inputs, so we omit their zero standard deviations.
Hyperparameter details and model-specific configurations are provided in Appendix~\ref{sec:appendix-hyperparameters} to support reproducibility.

\subsection{Overall Performance}
\label{sec:analysis}


We compare LightCRS with three categories of baselines: task-specific CRS methods, generative methods, and retrieval methods. 
For task-specific CRS methods, generative methods and our method, we rely on candidate-based recommendation setup with 10 items, using SBERT which shows the highest performance among retrievals methods~\cite{jeon2025adapting}.
The generative baselines are evaluated under
two input configurations: using item titles only (Vicuna-v1.5,
LLaVA-v1.5, and
LLaVA-v1.6) and using both item titles and product images
(LLaVA-v1.6, MiniMax-M2.7, and MiniMax-M3).
Table~\ref{tab:vram} compares LightCRS with all baselines in terms of average
recommendation performance, training cost, and inference efficiency. Since
the multimodal task-specific CRS methods are the only baselines that remain
competitive in recommendation accuracy, we restrict the detailed training-
and inference-cost comparison to this group. The complete per-domain
comparison is divided between two tables.
Table~\ref{tab:main-results} in
Appendix~\ref{sec:appendix-per-domain} reports LightCRS model with the task-specific CRSs baselines, while
Table~\ref{tab:zeroshot-baselines} in
Appendix~\ref{sec:appendix-unseen} provides the comparison with
the retrieval and generative baselines. Considered jointly, the two
tables compare LightCRS model against all baseline groups.
Table~\ref{tab:zeroshot-baselines} additionally reports the performance of LightCRS on unseen domains, which is further analyzed in Section~\ref{sec:further-analysis}.

\subsubsection{Comparison with task-specific CRSs baselines.}
Table~\ref{tab:main-results} compares LightCRS with the non-visual
task-specific methods RTA and
NBCRS, as well as the multimodal methods
LaViC, MSCRS, and
PRIME. Across the 25 domains, LightCRS achieves the best
performance in 24 domains across HR@1, NDCG@5, and MRR.
The results show a gap between the non-visual and multimodal
task-specific baselines. RTA achieves an average HR@1 of 0.0812, exceeding
the generative and retrieval baselines but remaining well below
the multimodal subgroup. Its performance also varies considerably across
domains, reaching an HR@1 of 0.2183 on Software and 0.2091 on Handmade,
both above its overall average and competitive with several
multimodal methods in these domains.

Among the multimodal baselines, PRIME achieves the strongest average results,
with an HR@1 of 0.1934, an NDCG@5 of 0.3798, and an MRR of 0.3680. The
results also show that the approximately 7.6B-parameter LaViC underperforms
the considerably smaller MSCRS (0.3B) and PRIME (2.2B), suggesting that
increasing model scale alone does not guarantee higher recommendation
performance.
In contrast, LightCRS achieves the best average results across all three
metrics, with an HR@1 of 0.2618, an NDCG@5 of 0.4540, and an MRR of
0.4367. Compared with PRIME, LightCRS improves HR@1 by 35.4\%, NDCG@5 by
19.5\%, and MRR by 18.7\%. The larger gain in HR@1 indicates that LightCRS
is particularly effective at identifying the single best item from the
candidate set, while its consistent improvements in NDCG@5 and MRR show
that it also produces a stronger overall candidate ranking.
LightCRS achieves these performance gains with  lower training
and inference costs. As reported in Table~\ref{tab:vram}, it requires only
5.3M trainable parameters, 0.1B total parameters, a 0.3\,GB model size,
0.3\,GB peak GPU memory, 97K MFLOPs per query, and 9.8\,ms per-query
latency, while supporting a throughput of 101.6 queries/s. Compared with
PRIME, LightCRS uses $1.7\times$ fewer trainable parameters and
$22\times$ fewer total parameters, while reducing model size by
$15\times$, peak GPU memory by $30\times$, per-query FLOPs by
$153\times$, latency by $266\times$, and increasing throughput by 254×. These results demonstrate that LightCRS
provides a stronger balance between recommendation accuracy and computational
efficiency than existing task-specific CRSs baselines.

\subsubsection{Comparison with retrieval and generative baselines.}
Combining LightCRS results reported in
Table~\ref{tab:main-results} with the baseline results reported in
Table~\ref{tab:zeroshot-baselines}, we compare LightCRS with the
retrieval methods BM25,
SBERT, RoBERTa, and
SimCSE; the generative models
Vicuna-v1.5,
LLaVA-v1.5, and
LLaVA-v1.6; and the proprietary models
MiniMax-M2.7 and MiniMax-M3.
The retrieval methods achieve average HR@1 values ranging from 0.0142
for BM25 to 0.0603 for SBERT. The open-source generative models achieve only
0.0629--0.0753 average HR@1 despite relying on larger
6.7--7.1B-parameter backbones, yielding only limited improvements over
SBERT. 
The results of both groups vary considerably across domains. For example,
SBERT reaches an HR@1 of 0.2166, an NDCG@5 of 0.4017, and an MRR of
0.3700 on Home. Its HR@1 in this domain surpasses every open-source
generative baseline, as well as RTA and MSCRS. LLaVA-v1.6 with image inputs achieves an HR@1 of 0.2227 on Software, exceeding its overall average performance. This indicates that incorporating product images into LLaVA-v1.6 has inconsistent effects across domains, improving recommendation performance in some domains while degrading it in others. This observation aligns with previous findings that directly incorporating candidate images does not necessarily improve recommendation accuracy~\cite{jeon2025adapting,liu2024rec}.
These domain-specific spikes show that textual
retrieval and general-purpose large model prompting can be effective in particular
categories, but their advantages do not transfer consistently across the
full set of domains.

The proprietary models perform better on average than the open-source generative baselines, with MiniMax-M2.7 and MiniMax-M3 achieving average
HR@1 values of 0.1025 and 0.1085, respectively. However, their results are
also strongly domain-dependent. MiniMax-M3 reaches an HR@1 of 0.3325 on
Home, giving it the second-highest baseline result in that domain, but
achieves only 0.1025 on Baby Products, where it falls behind
the multimodal task-specific methods and LightCRS.
In contrast, LightCRS achieves an average HR@1 of 0.2618, outperforming both
retrieval and generative models. At the per-domain level,
LightCRS achieves the highest HR@1 in 24 of the 25 domains against 
open-source baselines.
It also achieves the highest NDCG@5
and MRR in all 25 domains against the retrieval methods; these two
metrics are not reported for the generative baselines because they output
only a single recommendation rather than a candidate ranking.
Taken together, these results show that neither textual-similarity retrieval nor prompt-based scoring with general-purpose large models can achieve consistent recommendation performance across diverse product domains. In contrast, LightCRS leverages joint multi-domain training and multimodal product evidence, leading to robust performance across domains.

\begin{table*}[t]
\caption{Average recommendation performance and resource usage across methods. \protect\colorbox{gray!20}{Gray}-shaded rows denote multimodal task-specific CRS methods, training and inference costs for our method are compared only with those of methods in this group. ``--'' indicates that the value cannot be calculated. \textbf{Bold} and \underline{underlined} values denote the best and second-best results. Abbreviations: Rec. Perf.=recommendation performance, Train. P.=trainable parameters, Total P.=total parameters, GPU Mem.=peak GPU memory, Lat.=latency (ms/query), and Thpt.=throughput (queries/s).}
\label{tab:vram}
\centering
\footnotesize
\renewcommand{\arraystretch}{1.05}
\setlength{\tabcolsep}{3pt}

\begin{tabular}{l ccc ccc cccc}
\toprule
\multirow{2}{*}{Method}
& \multicolumn{3}{c}{Rec. Perf.}
& \multicolumn{3}{c}{Training Cost}
& \multicolumn{4}{c}{Inference Efficiency} \\
\cmidrule(lr){2-4} \cmidrule(lr){5-7} \cmidrule(lr){8-11}
& HR@1 $\uparrow$
& NDCG@5 $\uparrow$
& MRR $\uparrow$
& Train. P. $\downarrow$
& Total P. $\downarrow$
& Size $\downarrow$
& GPU Mem. $\downarrow$
& MFLOPs $\downarrow$
& Lat. $\downarrow$
& Thpt. $\uparrow$ \\
\midrule
\multicolumn{11}{l}{\textit{Retrieval}} \\
BM25~\cite{robertson2009probabilistic} & 0.0142 & 0.0434 & 0.1724 & -- & -- & 8.4MB & 0.0GB & 0.0 & 65.7 & 15.21 \\
SBERT~\cite{reimers2019sentence} & 0.0603 & 0.1224 & 0.2282 & -- & 0.1B & 0.4GB & 0.5GB & 70,675.1 & 17.5 & 58.14 \\
RoBERTa~\cite{liu2019roberta} & 0.0483 & 0.1065 & 0.2164 & -- & 0.4B & 1.4GB & 1.5GB & 154,634.3 & 24.5 & 40.94 \\
SimCSE~\cite{gao2021simcse} & 0.0195 & 0.0567 & 0.1837 & -- & 0.4B & 1.4GB & 1.5GB & 295,071.7 & 29.1 & 34.72 \\
\cmidrule(lr){1-11}
\multicolumn{11}{l}{\textit{Generative}} \\
Vicuna-v1.5~\cite{chiang2023vicuna} & 0.0633$_{\pm 0.0083}$ & -- & -- & -- & 6.7B & 13.5GB & 14.6GB & 28,435,892.3 & 1825.4 & 0.55 \\
LLaVA-v1.5~\cite{liu2024improved} & 0.0629$_{\pm 0.0112}$ & -- & -- & -- & 7.1B & 14.1GB & 15.3GB & 32,148,273.5 & 2087.6 & 0.48 \\
LLaVA-v1.6~\cite{liu2024llavanext} & 0.0683$_{\pm 0.0108}$ & -- & -- & -- & 7.1B & 14.1GB & 15.8GB & 35,832,914.7 & 2341.2 & 0.43 \\
LLaVA-v1.6 + title/image~\cite{liu2024llavanext} & 0.0753$_{\pm 0.0098}$ & -- & -- & -- & 7.1B & 14.1GB & 16.1GB & 39,527,168.8 & 2589.5 & 0.39 \\
MiniMax-M2.7 & 0.1025$_{\pm 0.0140}$ & -- & -- & -- & -- & -- & -- & -- & -- & -- \\
MiniMax-M3 & 0.1085$_{\pm 0.0143}$ & -- & -- & -- & -- & -- & -- & -- & -- & -- \\
\cmidrule(lr){1-11}
\multicolumn{11}{l}{\textit{Task-specific CRSs}} \\
NBCRS~\cite{xie2024neighborhood} & 0.0490$_{\pm 0.0118}$ & 0.0941$_{\pm 0.0120}$ & 0.1473$_{\pm 0.0097}$ & 22.9M & 0.02B & 0.1GB & 0.2GB & 7,344.7 & 3.2 & 319.4 \\
RTA~\cite{he2024reindex} & 0.0812$_{\pm 0.0074}$ & -- & -- & 40.3M & 6.8B & 13.6GB & 14.1GB & 6,352,337.1 & 116.8 & 8.57 \\
\rowcolor{gray!20} LaViC~\cite{jeon2025adapting} & 0.1338$_{\pm 0.0117}$ & -- & -- & 24.6M & 7.6B & 15.2GB & 16.2GB & 37,765,450.8 & 1940.3 & 0.5 \\
\rowcolor{gray!20} MSCRS~\cite{wei2025mscrs} & 0.1582$_{\pm 0.0077}$ & 0.2998$_{\pm 0.0078}$ & 0.2856$_{\pm 0.0069}$ & 22.2M & \underline{0.3B} & \underline{1.2GB} & \underline{2.0GB} & \underline{130,642.0} & \underline{112.2} & \underline{8.9} \\
\rowcolor{gray!20} PRIME~\cite{yue2025preference} & \underline{0.1934}$_{\pm 0.0143}$ & \underline{0.3798}$_{\pm 0.0117}$ & \underline{0.3680}$_{\pm 0.0111}$ & \underline{9.2M} & 2.2B & 4.4GB & 8.9GB & 14,849,125.6 & 2609.1 & 0.4 \\
\midrule 
LightCRS & \textbf{0.2618}$_{\pm 0.0104}$ & \textbf{0.4540}$_{\pm 0.0093}$ & \textbf{0.4367}$_{\pm 0.0083}$ & \textbf{5.3M} & \textbf{0.1B} & \textbf{0.3GB} & \textbf{0.3GB} & \textbf{97,078.6} & \textbf{9.8} & \textbf{101.6} \\
\bottomrule
\end{tabular}
\end{table*}

\subsection{Ablation Study}
\subsubsection{Component Ablation.}


To understand the contribution of each component in our framework, we conduct ablation experiments by removing one component at a time and evaluating the resulting recommendation performance, as shown in Table~\ref{tab:ablation}.

\paragraph{Inter-candidate comparison drives top-1 accuracy.}
Although removing inter-candidate comparison (ICC) retains the strongest performance among all ablated variants (HR@1 of 0.1432, NDCG@5 of 0.3505, MRR of 0.3389), it still reduces HR@1 by 45.3\% relative to the full model, confirming that direct candidate-to-candidate comparison is essential for accurate top-1 selection.

\paragraph{Cross-attention and visual tokens contribute most.}
Removing cross-attention (CA) or visual tokens (VT) has the greatest impact on top-1 accuracy, with HR@1 dropping to 0.0998 and 0.0987, respectively, underscoring the importance of query-conditioned multimodal interaction and visually grounded representations. Image descriptions (ID) further supply useful visual semantics, though their contribution is smaller than that of the raw visual tokens.

\paragraph{Title is the most informative textual view.}
Among the textual feature views, removing the title yields the lowest NDCG@5 (0.2832) and MRR (0.2871) among all ablated variants, while product descriptions and features contribute complementary signals.

\begin{table}[t]
\caption{Per-component ablation. Each row removes a single component from the full model, and we report the resulting average HR@1, NDCG@5, and MRR across all 25 product domains. Abbreviations: ICC~--~Inter-Candidate Comparison; ID~--~Image Descriptions; Feat~--~Features; Desc~--~Description; CA~--~Cross-Attention; VT~--~Visual Tokens.}
\label{tab:ablation}
\setlength{\tabcolsep}{4pt}
\begin{tabular}{lccc}
\toprule
Model Variant & HR@1 $\uparrow$ & NDCG@5 $\uparrow$ & MRR $\uparrow$ \\
\midrule
w/o ICC & 0.1432$_{\pm 0.0079}$ & 0.3505$_{\pm 0.0107}$ & 0.3389$_{\pm 0.0089}$ \\
w/o ID    & 0.1109$_{\pm 0.0069}$ & 0.3004$_{\pm 0.0034}$ & 0.3015$_{\pm 0.0039}$ \\
w/o Feat              & 0.1054$_{\pm 0.0064}$ & 0.2994$_{\pm 0.0054}$ & 0.2969$_{\pm 0.0047}$ \\
w/o Desc           & 0.1021$_{\pm 0.0002}$ & 0.3011$_{\pm 0.0017}$ & 0.2972$_{\pm 0.0006}$ \\
w/o Title                 & 0.1002$_{\pm 0.0012}$ & 0.2832$_{\pm 0.0023}$ & 0.2871$_{\pm 0.0015}$ \\
w/o CA       & 0.0998$_{\pm 0.0036}$ & 0.3052$_{\pm 0.0041}$ & 0.2971$_{\pm 0.0001}$ \\
w/o VT     & 0.0987$_{\pm 0.0024}$ & 0.2938$_{\pm 0.0033}$ & 0.2907$_{\pm 0.0002}$ \\
LightCRS (Ours) & \textbf{0.2618}$_{\pm 0.0104}$ & \textbf{0.4540}$_{\pm 0.0093}$ & \textbf{0.4367}$_{\pm 0.0083}$ \\
\bottomrule
\end{tabular}
\end{table}


\begin{table}[t]
\caption{Single-domain vs.\ Multi-domain. The single-domain regime trains a separate model per product category and we average metrics across the 25 domains; the full model trains one shared model jointly on all 25 domains. Abbreviations: Sing.=Single-domain, Multi.=Multi-domain.}
\label{tab:domain-fusion}
\setlength{\tabcolsep}{4pt}
\begin{tabular}{lccc}
\toprule
Training Regime & HR@1 $\uparrow$ & NDCG@5 $\uparrow$ & MRR $\uparrow$ \\
\midrule
Sing.         & 0.2272$_{\pm 0.0090}$ & 0.4305$_{\pm 0.0030}$ & 0.4113$_{\pm 0.0040}$ \\
Multi. (Ours)  & \textbf{0.2618}$_{\pm 0.0104}$ & \textbf{0.4540}$_{\pm 0.0093}$ & \textbf{0.4367}$_{\pm 0.0083}$ \\
\bottomrule
\end{tabular}
\end{table}

\subsubsection{Single-domain vs Multi-domain.}
Table~\ref{tab:domain-fusion} compares our full multi-domain model with a variant trained separately on each domain. In the single-domain setting, a separate model is trained for each product category, and HR@1, NDCG@5, and MRR are averaged across the 25 domain-specific models. Compared with this setting, our full multi-domain model improves HR@1, NDCG@5, and MRR by 15.2\%, 5.5\%, and 6.2\%, respectively, demonstrating that multi-domain knowledge sharing improves recommendation performance.

\section{Further Analysis}
\label{sec:further-analysis}


CRSs on unseen domains remains challenging, as even strong proprietary models achieve limited performance without task-specific adaptation~\cite{jeon2025adapting, xi2024memocrs, huang2025recommender, wang2024recmind, liao2026eliminating}. 
A key challenge in unseen-domain CRSs is the domain-dependent semantics. For example, ``light'' may denote low weight in hardware, light wearing comfort, a thin texture in cosmetics, or low computational cost in software.
Although LightCRS is not explicitly designed for recommendation on unseen domains, we evaluate its transferability through 5-fold domain-split testing across 25 domains. In each fold, the model is trained on 20 domains and evaluated on 5 unseen domains.
Table~\ref{tab:zeroshot-baselines} in Appendix~\ref{sec:appendix-unseen} shows the comparison of LightCRS on unseen domains with retrieval and generative baselines, for which the evaluation domains are also unseen.
LightCRS slightly outperforms the strongest MiniMax-M3 baseline. However, the absolute performance remains relatively low, indicating that recommendation on unseen domains remains challenging.

\section{Related Work}

Recommendation systems have long studied how to infer user preferences and rank relevant items from behavioral signals. Conventional sequential recommendation focuses on predicting users' next items from historical interaction sequences. Early methods model sequential transitions with Markov chains and recurrent networks~\cite{rendle2010factorizing,hidasi2015session}, followed by convolutional and self-attention architectures such as Caser~\cite{tang2018personalized}, SASRec~\cite{kang2018self}, and BERT4Rec~\cite{sun2019bert4rec}. More recent self-supervised and contrastive learning methods further improve sequence representations by exploiting auxiliary sequence-level signals~\cite{zhou2020s3,xie2022contrastive}. Along another line, candidate reranking methods refine an initially retrieved item list by modeling listwise context and candidate interactions. Representative methods include listwise context models~\cite{ai2018learning}, groupwise scoring functions~\cite{ai2019learning}, SetRank~\cite{pang2020setrank}, and personalized reranking modules~\cite{pei2019personalized}. Recent industrial generative recommender systems, such as OneMall~\cite{zhang2026onemall}, further reformulate large-scale e-commerce recommendation as autoregressive semantic-ID generation. Together, these studies provide important foundations for sequence modeling, candidate comparison, and industrial-scale recommendation, but they mainly rely on behavioral histories, retrieved item lists, or platform-specific interaction signals rather than natural-language conversational interactions.

CRSs extend recommendation by eliciting user preferences through multi-turn dialogue and refining recommendations according to the evolving conversational context. Early CRSs datasets and systems, such as ReDial~\cite{li2018towards}, GoRecDial~\cite{kang2019recommendation}, and TG-ReDial~\cite{zhou2020towards}, established conversational recommendation as the task of predicting relevant items from natural-language interactions, mostly in narrow domains such as movies. With the development of neural approaches, CRSs began to jointly model dialogue context, user preference, external knowledge, and item recommendation~\cite{chen2019towards,zhou2020improving,wang2022towards}. More recently, LLM-based CRSs methods have enhanced conversational recommendation by leveraging the instruction-following ability and world knowledge of large language models~\cite{he2023large,huang2025recommender,he2024reindex,zhu2025rank}. Despite these advances, most existing CRSs methods remain primarily text-based and are not designed to exploit rich multimodal product evidence in e-commerce scenarios.

Recent studies have further extended CRSs from text-only interactions to multimodal product understanding by leveraging VLMs to jointly encode visual and textual product evidence~\cite{jeon2025adapting,zhang2026moon,giahi2025vl}. These methods show that visual signals can complement textual dialogue, especially in visually grounded e-commerce scenarios where product appearance, style, color, and packaging are important for user decisions. However, VLMs-based CRSs still face three major limitations when deployed across diverse e-commerce categories: (1) training and inference costs that scale approximately linearly with the number of product domains, (2) performance degradation when combining heterogeneous domains into a single VLM, and (3) an additional routing challenge when multiple domain-specific models must be selected per dialogue. These limitations motivate LightCRS, which precomputes multimodal product views offline and performs lightweight dialogue-conditioned reranking online, using query-conditioned cross-attention and inter-candidate comparison to improve multi-domain multimodal CRSs recommendation accuracy while maintaining low training and inference costs.

\section{Conclusion}

We presented LightCRS, a lightweight framework for unified multi-domain multimodal conversational recommendation. By moving product representation computation offline, LightCRS avoids repeatedly encoding static candidate information during training and inference. Its compact online design enables a single model to adapt to different dialogue intents, share knowledge across product domains, and compare candidates jointly.  In addition, we introduce MM-R2A-25, a benchmark linking Reddit recommendation conversations to Amazon product listings across 25 categories. Experiments show that LightCRS provides state-of-the-art average recommendation performance compared to the evaluated baselines while using fewer trainable parameters and lower inference costs.

\paragraph{Limitations and Future Work.}
Our method follows a candidate-based reranking pipeline, so its overall performance remains partly constrained by the quality of the first-stage retriever. Improving candidate retrieval could therefore yield further gains in recommendation performance. 
On unseen domains, LightCRS still underperforms its in-domain supervised counterpart, while even strong proprietary models achieve limited performance. This indicates that achieving effective recommendation on unseen domains remains an important direction for future work.
Moreover, our model focuses on improving recommendation accuracy, while extending it to support dialogue generation remains another important direction for future work.



\bibliographystyle{ACM-Reference-Format}
\bibliography{sample-base}

\appendix

\section{Dataset Details}
\label{sec:appendix-dataset}

Table~\ref{tab:dataset} presents the full per-category breakdown of the \textit{MM-R2A-25} dataset. Table~\ref{tab:dataset-comparison} compares it with existing CRSs benchmarks. Early efforts (e.g., ReDial~\cite{li2018towards}, GoRecDial~\cite{kang2019recommendation}, and TG-ReDial~\cite{zhou2020towards}) primarily focus on movies, often with crowd-sourced or synthetic dialogues. Recent datasets have expanded into broader domains (e.g., beauty, home and fashion) or used natural Reddit conversations~\cite{rabiah2024bridging, jia2022convrec, liu2023u}, but still lack multimodal product information. In contrast, \textit{MM-R2A-25} covers 25 diverse product categories, enabling comprehensive evaluations of multi-domain multimodal conversational recommender systems. Following LaViC~\cite{jeon2025adapting}, we use SBERT to retrieve candidate sets for each conversation, with a Recall@10 of 0.6479.

\begin{table*}
  \caption{Summary of the MM-R2A-25 dataset. \#Conv: number of conversations. \#Turns: number of conversational turns. \#Items: number of unique products. \#Images: total number of verified product images.}
  \label{tab:dataset}
  \begin{tabular}{lrrrr}
    \toprule
    Category & \#Conv & \#Turns & \#Items & \#Images \\
    \midrule
    Appliances               & 2,291   & 4,582   & 739     & 4,062   \\
    Arts, Crafts \& Sewing   & 1,267   & 2,315   & 823     & 3,473   \\
    Automotive               & 1,512   & 2,869   & 811     & 3,657   \\
    Baby Products            & 3,336   & 6,672   & 1,975   & 11,561  \\
    Beauty                   & 7,672   & 22,966  & 5,433   & 28,082  \\
    Books                    & 4,707   & 13,250  & 1,978   & 1,864   \\
    CDs \& Vinyl             & 3,818   & 7,345   & 1,023   & 1,562   \\
    Cell Phones \& Accessories & 3,576 & 7,612   & 1,926   & 11,580  \\
    Digital Music            & 1,350   & 2,700   & 613     & 988     \\
    Electronics              & 3,996   & 7,992   & 2,913   & 12,323  \\
    Fashion                  & 8,039   & 21,831  & 6,716   & 31,162  \\
    Grocery \& Gourmet Food  & 2,864   & 5,222   & 1,940   & 8,186   \\
    Handmade Products        & 2,856   & 5,712   & 781     & 4,672   \\
    Health                   & 1,590   & 2,974   & 1,134   & 6,050   \\
    Industrial \& Scientific & 3,978   & 7,411   & 2,268   & 10,520  \\
    Home                     & 3,701   & 6,675   & 3,077   & 18,505 \\
    Movies \& TV             & 2,284   & 4,455   & 972     & 1,315   \\
    Musical Instruments       & 2,230   & 4,460   & 1,327   & 4,693   \\
    Office Products          & 2,905   & 5,513   & 1,684   & 8,375   \\
    Patio, Lawn \& Garden    & 1,148   & 2,044   & 874     & 3,989   \\
    Pet Supplies             & 5,219   & 9,628   & 2,708   & 13,293  \\
    Software                 & 2,277   & 4,554   & 283     & 1,625   \\
    Sports \& Outdoors       & 3,180   & 7,405   & 1,156   & 6,459   \\
    Toys \& Games            & 1,909   & 3,714   & 1,113   & 5,462   \\
    Video Games              & 19,142  & 38,284  & 5,854   & 15,810  \\
    \bottomrule
  \end{tabular}
\end{table*}

\begin{table*}
  \caption{Comparison of existing conversational recommendation datasets. \textit{MM-R2A-25} consists of natural Reddit conversations aligned with Amazon Reviews 2023 across 25 diverse product categories.}
  \label{tab:dataset-comparison}
  \begin{tabular}{lrrlll}
    \toprule
    Dataset & \#Conv & \#Turns & \#Items & Domain & Source \\
    \midrule
    Facebook Rec~\cite{dodge2015evaluating} & 1M & 6M & -- & Movies & Synthetic \\
    ReDial~\cite{li2018towards} & 10K & 182K & 6.2K & Movies & Crowd-sourced \\
    GoRecDial~\cite{kang2019recommendation} & 9K & 170K & -- & Movies & Crowd-sourced \\
    OpenDialKG~\cite{moon2019opendialkg} & 15K & 91K & -- & Movies, Music, etc. & Crowd-sourced \\
    TG-ReDial~\cite{zhou2020towards} & 10K & 129K & -- & Movies & Synthetic \\
    DuRecDial 2.0~\cite{liu2021durecdial} & 16.5K & 255K & -- & Movies, Music, etc. & Crowd-sourced \\
    CCPE-M~\cite{radlinski2019coached} & 502 & 11K & -- & Movies & Crowd-sourced \\
    INSPIRED~\cite{hayati2020inspired} & 1K & 35K & 1.9K & Movies & Crowd-sourced \\
    Reddit-Movie Base~\cite{rabiah2024bridging} & 85K & 133K & 24.3K & Movies & Natural \\
    Reddit-Movie Large~\cite{rabiah2024bridging} & 634K & 1.6M & 51.2K & Movies & Natural \\
    U-Need~\cite{liu2023u} & 7K & 53K & -- & E-commerce & Natural \\
    E-ConvRec~\cite{jia2022convrec} & 25K & 775K & -- & E-commerce & Natural \\
    Hoops~\cite{fu2021hoops} & -- & 11.6M & -- & E-commerce & Synthetic \\
    MGConvRec~\cite{xu2020user} & 7K & 73K & -- & Restaurant & Crowd-sourced \\
    MMConv~\cite{liao2021mmconv} & 5K & 39K & -- & Travel & Crowd-sourced \\
    MobileConvRec~\cite{maji2024mobileconvrec} & 12.2K & 156K & 1.7K & Mobile Apps & Synthetic \\
        Pearl~\cite{kim2024pearl} & 57.2K & 482K & 9.4K & Movies & Synthetic \\
    LLM-REDIAL~\cite{liang2024llm} & 47.6K & 482.6K & -- & Movie, Book, Sports & Synthetic \\
    MUSE~\cite{wang2025muse} & 7K & 83K & 13.7K & Clothing, Shoes, Jewelry & Synthetic \\
    SCREEN~\cite{lin2024screen} & 20K & -- & -- & Movies & Synthetic \\
    Reddit-Amazon~\cite{jeon2025adapting} & 19K & 51K & 15K & Beauty, Fashion, Home & Natural \\
    Marqo-GS-10M~\cite{zhu2025generalized} & 98.2K & - & 5.50M & Fashion, Home & Natural \\
    \cmidrule(lr){1-6}
    \textbf{MM-R2A-25 (Ours)} & \textbf{97K} & \textbf{208K} & \textbf{50K} & \textbf{Multi-domain} & \textbf{Natural} \\
    \bottomrule
  \end{tabular}
\end{table*}

\section{Per-Domain Results}
\label{sec:appendix-per-domain}

Table~\ref{tab:main-results} reports the full per-domain breakdown of HR@1, NDCG@5, and MRR across all 25 domains in \textit{MM-R2A-25}, comparing LightCRS with three baseline groups: task-specific CRS, generative, retrieval methods. These results complement the aggregate analysis in Section~\ref{sec:analysis}.

\begin{table*}[t]
\caption{Per-domain comparison of the task-specific CRS baselines and LightCRS across 25 Amazon product categories. Panel~A reports HR@1 for all methods, whereas Panel~B reports NDCG@5 and MRR only for ranking-based methods. \textbf{Bold} and \underline{underline} indicate the best and second-best results, respectively.}
\label{tab:main-results}
\centering
\scriptsize

\textbf{Panel A: HR@1.}

\begin{tabular}{l cccccc}
\toprule
& \multicolumn{6}{c}{HR@1 $\uparrow$} \\
\cmidrule(lr){2-7}
& \multicolumn{5}{c}{Task-specific CRSs} & \multicolumn{1}{c}{Ours} \\
\cmidrule(lr){2-6} \cmidrule(lr){7-7}
Domain & NBCRS~\cite{xie2024neighborhood} & RTA~\cite{he2024reindex} & LaViC~\cite{jeon2025adapting} & MSCRS~\cite{wei2025mscrs} & PRIME~\cite{yue2025preference} & LightCRS \\
\midrule
Appliances & 0.0957$_{\pm 0.0101}$ & 0.2130$_{\pm 0.0090}$ & \underline{0.2709}$_{\pm 0.0014}$ & 0.2391$_{\pm 0.0063}$ & 0.2261$_{\pm 0.0223}$ & \textbf{0.2873}$_{\pm 0.0152}$ \\
Arts \& Crafts & 0.0078$_{\pm 0.0091}$ & 0.0469$_{\pm 0.0048}$ & 0.1013$_{\pm 0.0031}$ & 0.0859$_{\pm 0.0059}$ & \underline{0.1016}$_{\pm 0.0036}$ & \textbf{0.1349}$_{\pm 0.0270}$ \\
Automotive & 0.0197$_{\pm 0.0090}$ & 0.0395$_{\pm 0.0051}$ & 0.0560$_{\pm 0.0253}$ & 0.0526$_{\pm 0.0053}$ & \underline{0.0852}$_{\pm 0.0128}$ & \textbf{0.1184}$_{\pm 0.0139}$ \\
Baby Products & 0.0567$_{\pm 0.0190}$ & 0.0090$_{\pm 0.0058}$ & 0.3373$_{\pm 0.0113}$ & 0.5254$_{\pm 0.0168}$ & \underline{0.5791}$_{\pm 0.0136}$ & \textbf{0.7194}$_{\pm 0.0093}$ \\
Beauty & 0.0326$_{\pm 0.0137}$ & 0.0618$_{\pm 0.0069}$ & 0.1187$_{\pm 0.0224}$ & 0.1135$_{\pm 0.0050}$ & \underline{0.2317}$_{\pm 0.0122}$ & \textbf{0.4009}$_{\pm 0.0106}$ \\
Books & 0.0275$_{\pm 0.0065}$ & 0.0678$_{\pm 0.0044}$ & 0.0758$_{\pm 0.0032}$ & 0.0742$_{\pm 0.0093}$ & \underline{0.0994}$_{\pm 0.0025}$ & \textbf{0.1081}$_{\pm 0.0061}$ \\
CDs \& Vinyl & 0.0470$_{\pm 0.0091}$ & 0.0940$_{\pm 0.0096}$ & 0.0887$_{\pm 0.0031}$ & \underline{0.1017}$_{\pm 0.0025}$ & 0.0877$_{\pm 0.0045}$ & \textbf{0.1048}$_{\pm 0.0129}$ \\
Cell Phones & 0.0279$_{\pm 0.0144}$ & 0.0641$_{\pm 0.0060}$ & 0.1002$_{\pm 0.0077}$ & 0.0752$_{\pm 0.0048}$ & \underline{0.1170}$_{\pm 0.0051}$ & \textbf{0.1225}$_{\pm 0.0036}$ \\
Digital Music & 0.1149$_{\pm 0.0101}$ & 0.1897$_{\pm 0.0078}$ & \underline{0.2643}$_{\pm 0.0054}$ & 0.1782$_{\pm 0.0169}$ & 0.1842$_{\pm 0.0025}$ & \textbf{0.2674}$_{\pm 0.0129}$ \\
Electronics & 0.0449$_{\pm 0.0083}$ & 0.1870$_{\pm 0.0011}$ & 0.0075$_{\pm 0.0080}$ & 0.2843$_{\pm 0.0193}$ & \underline{0.2868}$_{\pm 0.0174}$ & \textbf{0.5885}$_{\pm 0.0051}$ \\
Fashion & 0.0130$_{\pm 0.0043}$ & 0.0281$_{\pm 0.0011}$ & 0.1232$_{\pm 0.0056}$ & 0.0982$_{\pm 0.0104}$ & \underline{0.3276}$_{\pm 0.0207}$ & \textbf{0.3517}$_{\pm 0.0025}$ \\
Grocery & 0.0070$_{\pm 0.0102}$ & 0.0314$_{\pm 0.0075}$ & 0.0940$_{\pm 0.0165}$ & 0.1185$_{\pm 0.0073}$ & \underline{0.1209}$_{\pm 0.0114}$ & \textbf{0.1376}$_{\pm 0.0051}$ \\
Handmade & 0.1429$_{\pm 0.0219}$ & 0.2091$_{\pm 0.0112}$ & 0.2264$_{\pm 0.0028}$ & \underline{0.2976}$_{\pm 0.0003}$ & 0.2509$_{\pm 0.0302}$ & \textbf{0.3175}$_{\pm 0.0097}$ \\
Health & 0.0278$_{\pm 0.0146}$ & 0.0389$_{\pm 0.0074}$ & 0.0777$_{\pm 0.0128}$ & 0.0722$_{\pm 0.0163}$ & \underline{0.1160}$_{\pm 0.0155}$ & \textbf{0.1389}$_{\pm 0.0060}$ \\
Home & 0.0252$_{\pm 0.0183}$ & 0.0579$_{\pm 0.0050}$ & \underline{0.3197}$_{\pm 0.0025}$ & 0.0605$_{\pm 0.0066}$ & \textbf{0.3575}$_{\pm 0.0034}$ & 0.2002$_{\pm 0.0071}$ \\
Industrial & 0.0150$_{\pm 0.0123}$ & 0.0551$_{\pm 0.0099}$ & 0.0902$_{\pm 0.0195}$ & 0.0501$_{\pm 0.0043}$ & \underline{0.0982}$_{\pm 0.0287}$ & \textbf{0.1441}$_{\pm 0.0146}$ \\
Movies \& TV & 0.0480$_{\pm 0.0037}$ & 0.0655$_{\pm 0.0112}$ & 0.0786$_{\pm 0.0044}$ & 0.0655$_{\pm 0.0072}$ & \underline{0.1059}$_{\pm 0.0133}$ & \textbf{0.1376}$_{\pm 0.0158}$ \\
Musical Instruments & 0.1045$_{\pm 0.0118}$ & 0.0261$_{\pm 0.0064}$ & 0.0475$_{\pm 0.0229}$ & 0.3848$_{\pm 0.0096}$ & \underline{0.4821}$_{\pm 0.0259}$ & \textbf{0.6627}$_{\pm 0.0076}$ \\
Office Prod. & 0.0206$_{\pm 0.0065}$ & 0.0619$_{\pm 0.0053}$ & \underline{0.0846}$_{\pm 0.0042}$ & 0.0845$_{\pm 0.0048}$ & 0.0790$_{\pm 0.0067}$ & \textbf{0.0922}$_{\pm 0.0011}$ \\
Patio \& Garden & 0.0086$_{\pm 0.0090}$ & 0.0345$_{\pm 0.0051}$ & \underline{0.1206}$_{\pm 0.0022}$ & 0.0431$_{\pm 0.0032}$ & 0.0862$_{\pm 0.0266}$ & \textbf{0.1609}$_{\pm 0.0289}$ \\
Pet Supplies & 0.0287$_{\pm 0.0231}$ & 0.0593$_{\pm 0.0096}$ & 0.0792$_{\pm 0.0246}$ & 0.0899$_{\pm 0.0079}$ & \underline{0.0955}$_{\pm 0.0117}$ & \textbf{0.1128}$_{\pm 0.0058}$ \\
Software & 0.1659$_{\pm 0.0084}$ & 0.2183$_{\pm 0.0078}$ & 0.3249$_{\pm 0.0320}$ & \underline{0.3319}$_{\pm 0.0031}$ & 0.3057$_{\pm 0.0098}$ & \textbf{0.4607}$_{\pm 0.0244}$ \\
Sports & 0.0280$_{\pm 0.0115}$ & 0.0592$_{\pm 0.0111}$ & \underline{0.0903}$_{\pm 0.0134}$ & 0.0841$_{\pm 0.0056}$ & 0.0709$_{\pm 0.0322}$ & \textbf{0.1106}$_{\pm 0.0057}$ \\
Toys \& Games & 0.0104$_{\pm 0.0170}$ & 0.0417$_{\pm 0.0124}$ & 0.0833$_{\pm 0.0138}$ & 0.0625$_{\pm 0.0052}$ & \underline{0.1301}$_{\pm 0.0068}$ & \textbf{0.1422}$_{\pm 0.0051}$ \\
Video Games & 0.1055$_{\pm 0.0124}$ & 0.0700$_{\pm 0.0128}$ & 0.0851$_{\pm 0.0239}$ & \underline{0.3817}$_{\pm 0.0093}$ & 0.2089$_{\pm 0.0192}$ & \textbf{0.5240}$_{\pm 0.0045}$ \\
\cmidrule(lr){1-7}
\textbf{Avg.} & 0.0490$_{\pm 0.0118}$ & 0.0812$_{\pm 0.0074}$ & 0.1338$_{\pm 0.0117}$ & 0.1582$_{\pm 0.0077}$ & \underline{0.1934}$_{\pm 0.0143}$ & \textbf{0.2618}$_{\pm 0.0104}$ \\
\bottomrule
\end{tabular}

\vspace{1em}
\textbf{Panel B: NDCG@5 and MRR.}

\begin{tabular}{l cccc cccc}
\toprule
& \multicolumn{4}{c}{NDCG@5 $\uparrow$} & \multicolumn{4}{c}{MRR $\uparrow$} \\
\cmidrule(lr){2-5} \cmidrule(lr){6-9}
& \multicolumn{3}{c}{Task-specific CRSs} & \multicolumn{1}{c}{Ours} & \multicolumn{3}{c}{Task-specific CRSs} & \multicolumn{1}{c}{Ours} \\
\cmidrule(lr){2-4} \cmidrule(lr){5-5} \cmidrule(lr){6-8} \cmidrule(lr){9-9}
Domain & NBCRS~\cite{xie2024neighborhood} & MSCRS~\cite{wei2025mscrs} & PRIME~\cite{yue2025preference} & LightCRS & NBCRS~\cite{xie2024neighborhood} & MSCRS~\cite{wei2025mscrs} & PRIME~\cite{yue2025preference} & LightCRS \\
\midrule
Appliances & 0.1614$_{\pm 0.0112}$ & 0.4365$_{\pm 0.0068}$ & \underline{0.4437}$_{\pm 0.0048}$ & \textbf{0.5160}$_{\pm 0.0180}$ & 0.1861$_{\pm 0.0109}$ & 0.3925$_{\pm 0.0065}$ & \underline{0.4202}$_{\pm 0.0127}$ & \textbf{0.4781}$_{\pm 0.0140}$ \\
Arts \& Crafts & 0.0296$_{\pm 0.0051}$ & 0.2332$_{\pm 0.0045}$ & \underline{0.2810}$_{\pm 0.0054}$ & \textbf{0.3123}$_{\pm 0.0103}$ & 0.1070$_{\pm 0.0070}$ & 0.2170$_{\pm 0.0043}$ & \underline{0.2691}$_{\pm 0.0269}$ & \textbf{0.3130}$_{\pm 0.0161}$ \\
Automotive & 0.0544$_{\pm 0.0071}$ & 0.1812$_{\pm 0.0037}$ & \underline{0.2842}$_{\pm 0.0101}$ & \textbf{0.3070}$_{\pm 0.0048}$ & 0.1187$_{\pm 0.0062}$ & 0.1832$_{\pm 0.0032}$ & \underline{0.2726}$_{\pm 0.0027}$ & \textbf{0.3068}$_{\pm 0.0068}$ \\
Baby Products & 0.1065$_{\pm 0.0172}$ & 0.5908$_{\pm 0.0124}$ & \underline{0.6910}$_{\pm 0.0098}$ & \textbf{0.8339}$_{\pm 0.0069}$ & 0.1515$_{\pm 0.0196}$ & 0.5887$_{\pm 0.0143}$ & \underline{0.6713}$_{\pm 0.0088}$ & \textbf{0.8099}$_{\pm 0.0103}$ \\
Beauty & 0.0581$_{\pm 0.0104}$ & 0.2132$_{\pm 0.0046}$ & \underline{0.3814}$_{\pm 0.0135}$ & \textbf{0.6133}$_{\pm 0.0067}$ & 0.1295$_{\pm 0.0118}$ & 0.2008$_{\pm 0.0047}$ & \underline{0.3732}$_{\pm 0.0141}$ & \textbf{0.5767}$_{\pm 0.0083}$ \\
Books & 0.0682$_{\pm 0.0073}$ & 0.2381$_{\pm 0.0101}$ & \underline{0.2817}$_{\pm 0.0155}$ & \textbf{0.2984}$_{\pm 0.0080}$ & 0.1250$_{\pm 0.0059}$ & 0.2221$_{\pm 0.0088}$ & \underline{0.2817}$_{\pm 0.0163}$ & \textbf{0.2990}$_{\pm 0.0059}$ \\
CDs \& Vinyl & 0.1398$_{\pm 0.0103}$ & \underline{0.2730}$_{\pm 0.0040}$ & 0.2700$_{\pm 0.0031}$ & \textbf{0.2892}$_{\pm 0.0150}$ & 0.1443$_{\pm 0.0093}$ & 0.2584$_{\pm 0.0041}$ & \underline{0.2691}$_{\pm 0.0020}$ & \textbf{0.2861}$_{\pm 0.0125}$ \\
Cell Phones & 0.0792$_{\pm 0.0141}$ & 0.2295$_{\pm 0.0038}$ & \underline{0.3189}$_{\pm 0.0043}$ & \textbf{0.3203}$_{\pm 0.0044}$ & 0.1256$_{\pm 0.0077}$ & 0.2148$_{\pm 0.0023}$ & \underline{0.3133}$_{\pm 0.0078}$ & \textbf{0.3170}$_{\pm 0.0032}$ \\
Digital Music & 0.1696$_{\pm 0.0099}$ & 0.3693$_{\pm 0.0139}$ & \underline{0.4979}$_{\pm 0.0137}$ & \textbf{0.5104}$_{\pm 0.0093}$ & 0.2045$_{\pm 0.0077}$ & 0.3310$_{\pm 0.0116}$ & \underline{0.4704}$_{\pm 0.0013}$ & \textbf{0.4726}$_{\pm 0.0091}$ \\
Electronics & 0.0718$_{\pm 0.0051}$ & 0.3894$_{\pm 0.0136}$ & \underline{0.4346}$_{\pm 0.0315}$ & \textbf{0.7475}$_{\pm 0.0026}$ & 0.1424$_{\pm 0.0066}$ & 0.3781$_{\pm 0.0157}$ & \underline{0.4145}$_{\pm 0.0260}$ & \textbf{0.7129}$_{\pm 0.0041}$ \\
Fashion & 0.0266$_{\pm 0.0020}$ & 0.1558$_{\pm 0.0076}$ & \underline{0.4500}$_{\pm 0.0045}$ & \textbf{0.5745}$_{\pm 0.0021}$ & 0.1127$_{\pm 0.0023}$ & 0.1564$_{\pm 0.0053}$ & \underline{0.4278}$_{\pm 0.0073}$ & \textbf{0.5354}$_{\pm 0.0018}$ \\
Grocery & 0.0290$_{\pm 0.0267}$ & 0.2482$_{\pm 0.0102}$ & \underline{0.2926}$_{\pm 0.0116}$ & \textbf{0.3237}$_{\pm 0.0102}$ & 0.1067$_{\pm 0.0134}$ & 0.2395$_{\pm 0.0057}$ & \underline{0.2958}$_{\pm 0.0133}$ & \textbf{0.3206}$_{\pm 0.0031}$ \\
Handmade & 0.2208$_{\pm 0.0147}$ & 0.4762$_{\pm 0.0158}$ & \underline{0.5001}$_{\pm 0.0131}$ & \textbf{0.5217}$_{\pm 0.0010}$ & 0.2286$_{\pm 0.0158}$ & 0.4505$_{\pm 0.0180}$ & \underline{0.4591}$_{\pm 0.0101}$ & \textbf{0.4902}$_{\pm 0.0063}$ \\
Health & 0.0455$_{\pm 0.0132}$ & 0.2061$_{\pm 0.0166}$ & \underline{0.3392}$_{\pm 0.0207}$ & \textbf{0.3653}$_{\pm 0.0057}$ & 0.1255$_{\pm 0.0113}$ & 0.2010$_{\pm 0.0136}$ & \underline{0.3194}$_{\pm 0.0040}$ & \textbf{0.3302}$_{\pm 0.0052}$ \\
Home & 0.0453$_{\pm 0.0136}$ & 0.1278$_{\pm 0.0040}$ & \textbf{0.5073}$_{\pm 0.0076}$ & \underline{0.4408}$_{\pm 0.0044}$ & 0.1227$_{\pm 0.0137}$ & 0.1557$_{\pm 0.0044}$ & \textbf{0.4860}$_{\pm 0.0086}$ & \underline{0.4068}$_{\pm 0.0058}$ \\
Industrial & 0.0456$_{\pm 0.0127}$ & 0.1704$_{\pm 0.0042}$ & \underline{0.3087}$_{\pm 0.0123}$ & \textbf{0.3358}$_{\pm 0.0144}$ & 0.1145$_{\pm 0.0103}$ & 0.1666$_{\pm 0.0030}$ & \underline{0.2935}$_{\pm 0.0249}$ & \textbf{0.3311}$_{\pm 0.0108}$ \\
Movies \& TV & 0.0823$_{\pm 0.0030}$ & 0.2111$_{\pm 0.0055}$ & \underline{0.3013}$_{\pm 0.0020}$ & \textbf{0.3272}$_{\pm 0.0120}$ & 0.1433$_{\pm 0.0029}$ & 0.2001$_{\pm 0.0047}$ & \underline{0.3121}$_{\pm 0.0028}$ & \textbf{0.3239}$_{\pm 0.0098}$ \\
Musical Instruments & 0.1761$_{\pm 0.0122}$ & 0.5504$_{\pm 0.0080}$ & \underline{0.6189}$_{\pm 0.0038}$ & \textbf{0.7932}$_{\pm 0.0081}$ & 0.1942$_{\pm 0.0092}$ & 0.5182$_{\pm 0.0071}$ & \underline{0.6018}$_{\pm 0.0059}$ & \textbf{0.7663}$_{\pm 0.0061}$ \\
Office Prod. & 0.0498$_{\pm 0.0063}$ & 0.2369$_{\pm 0.0046}$ & \underline{0.2648}$_{\pm 0.0031}$ & \textbf{0.2760}$_{\pm 0.0059}$ & 0.1187$_{\pm 0.0053}$ & 0.2284$_{\pm 0.0049}$ & \underline{0.2623}$_{\pm 0.0109}$ & \textbf{0.2792}$_{\pm 0.0050}$ \\
Patio \& Garden & 0.0172$_{\pm 0.0109}$ & 0.1736$_{\pm 0.0031}$ & \underline{0.2875}$_{\pm 0.0146}$ & \textbf{0.3429}$_{\pm 0.0389}$ & 0.1079$_{\pm 0.0089}$ & 0.1702$_{\pm 0.0038}$ & \underline{0.2933}$_{\pm 0.0055}$ & \textbf{0.3448}$_{\pm 0.0354}$ \\
Pet Supplies & 0.0582$_{\pm 0.0183}$ & 0.2232$_{\pm 0.0045}$ & \underline{0.2813}$_{\pm 0.0213}$ & \textbf{0.3234}$_{\pm 0.0024}$ & 0.1259$_{\pm 0.0166}$ & 0.2159$_{\pm 0.0052}$ & \underline{0.2903}$_{\pm 0.0134}$ & \textbf{0.3145}$_{\pm 0.0027}$ \\
Software & 0.3367$_{\pm 0.0075}$ & \underline{0.5535}$_{\pm 0.0021}$ & 0.5477$_{\pm 0.0160}$ & \textbf{0.6819}$_{\pm 0.0147}$ & 0.2994$_{\pm 0.0059}$ & \underline{0.5149}$_{\pm 0.0029}$ & 0.5047$_{\pm 0.0121}$ & \textbf{0.6359}$_{\pm 0.0131}$ \\
Sports & 0.0877$_{\pm 0.0128}$ & 0.2283$_{\pm 0.0053}$ & \underline{0.2794}$_{\pm 0.0148}$ & \textbf{0.3005}$_{\pm 0.0064}$ & 0.1279$_{\pm 0.0072}$ & 0.2152$_{\pm 0.0035}$ & \underline{0.2766}$_{\pm 0.0171}$ & \textbf{0.2998}$_{\pm 0.0023}$ \\
Toys \& Games & 0.0311$_{\pm 0.0265}$ & 0.2179$_{\pm 0.0084}$ & \underline{0.2978}$_{\pm 0.0092}$ & \textbf{0.3228}$_{\pm 0.0157}$ & 0.1252$_{\pm 0.0168}$ & 0.1971$_{\pm 0.0054}$ & \underline{0.3013}$_{\pm 0.0079}$ & \textbf{0.3195}$_{\pm 0.0078}$ \\
Video Games & 0.1617$_{\pm 0.0210}$ & \underline{0.5617}$_{\pm 0.0188}$ & 0.3338$_{\pm 0.0264}$ & \textbf{0.6724}$_{\pm 0.0051}$ & 0.1955$_{\pm 0.0092}$ & \underline{0.5233}$_{\pm 0.0084}$ & 0.3296$_{\pm 0.0151}$ & \textbf{0.6483}$_{\pm 0.0031}$ \\
\cmidrule(lr){1-9}
\textbf{Avg.} & 0.0941$_{\pm 0.0120}$ & 0.2998$_{\pm 0.0078}$ & \underline{0.3798}$_{\pm 0.0117}$ & \textbf{0.4540}$_{\pm 0.0093}$ & 0.1473$_{\pm 0.0097}$ & 0.2856$_{\pm 0.0069}$ & \underline{0.3680}$_{\pm 0.0111}$ & \textbf{0.4367}$_{\pm 0.0083}$ \\
\bottomrule
\end{tabular}
\end{table*}

\section{Generalization to Held-Out Unseen Domains}
\label{sec:appendix-unseen}

To provide a supplementary assessment of unseen-domain generalization, we conduct a five-fold held-out-domain evaluation. The 25 product domains are partitioned into 5 non-overlapping groups, and in each fold, LightCRS is trained on 20 domains and evaluated on the remaining 5 domains, whose data are entirely excluded from training. Table~\ref{tab:zeroshot-baselines} compares LightCRS under this unseen-domain setting with retrieval and generative baselines, for which the evaluation domains are also unseen.

\begin{table*}[t]
\caption{Comparison of retrieval, generative baselines, and LightCRS on unseen domains across 25 Amazon product categories. Panel~A reports HR@1; Panel~B reports NDCG@5 and MRR for ranking-based methods. \textbf{Bold} and \underline{underline} indicate the best and second-best results within each evaluation setting. Abbreviations: RoB. denotes RoBERTa\textsubscript{large}; Vic. denotes Vicuna-v1.5; 1.5, 1.6, and 1.6+I denote LLaVA-v1.5, LLaVA-v1.6, and LLaVA-v1.6+image; M3 denotes MiniMax-M3; and M2.7 denotes MiniMax-M2.7.
}
\label{tab:zeroshot-baselines}
\centering
\scriptsize

\textbf{Panel A: HR@1.}

\begin{tabular}{l cccc cccc cc c}
\toprule
& \multicolumn{11}{c}{HR@1 $\uparrow$} \\
\cmidrule(lr){2-12}
& \multicolumn{4}{c}{Retrieval} & \multicolumn{6}{c}{Generative} & \multicolumn{1}{c}{Ours} \\
\cmidrule(lr){2-5} \cmidrule(lr){6-11} \cmidrule(lr){12-12}
Domain & BM25~\cite{robertson2009probabilistic} & SBERT~\cite{reimers2019sentence} & RoB.~\cite{liu2019roberta} & SimCSE~\cite{gao2021simcse} & Vic.~\cite{chiang2023vicuna} & 1.5~\cite{liu2024improved} & 1.6~\cite{liu2024llavanext} & 1.6+I~\cite{liu2024llavanext} & M2.7 & M3 & LightCRS \\
\midrule
Appliances & 0.0076 & 0.0684 & 0.0456 & 0.0000 & 0.0783$_{\pm 0.0024}$ & 0.0783$_{\pm 0.0059}$ & 0.0739$_{\pm 0.0133}$ & 0.1174$_{\pm 0.0105}$ & \underline{0.1957}$_{\pm 0.0033}$ & \textbf{0.2174}$_{\pm 0.0014}$ & 0.0994$_{\pm 0.0208}$ \\
Arts \& Crafts & 0.0000 & 0.0153 & 0.0230 & 0.0154 & 0.0547$_{\pm 0.0061}$ & 0.0546$_{\pm 0.0076}$ & 0.0551$_{\pm 0.0154}$ & 0.0560$_{\pm 0.0110}$ & 0.0703$_{\pm 0.0240}$ & \underline{0.0880}$_{\pm 0.0035}$ & \textbf{0.1198}$_{\pm 0.0252}$ \\
Automotive & 0.0000 & 0.0128 & 0.0132 & 0.0134 & 0.0197$_{\pm 0.0008}$ & 0.0395$_{\pm 0.0126}$ & 0.0197$_{\pm 0.0168}$ & 0.0526$_{\pm 0.0100}$ & \underline{0.0612}$_{\pm 0.0151}$ & 0.0329$_{\pm 0.0211}$ & \textbf{0.1096}$_{\pm 0.0312}$ \\
Baby Products & 0.0627 & 0.0657 & 0.0478 & 0.0209 & 0.1164$_{\pm 0.0128}$ & 0.1194$_{\pm 0.0162}$ & 0.1463$_{\pm 0.0156}$ & 0.0657$_{\pm 0.0042}$ & 0.1007$_{\pm 0.0220}$ & 0.1025$_{\pm 0.0174}$ & \textbf{0.1363}$_{\pm 0.0125}$ \\
Beauty & 0.0169 & 0.0551 & 0.0640 & 0.0326 & 0.0533$_{\pm 0.0131}$ & 0.0476$_{\pm 0.0165}$ & 0.0770$_{\pm 0.0168}$ & 0.0584$_{\pm 0.0106}$ & 0.1148$_{\pm 0.0219}$ & \underline{0.1237}$_{\pm 0.0132}$ & \textbf{0.1511}$_{\pm 0.0133}$ \\
Books & 0.0085 & 0.0486 & 0.0275 & 0.0042 & 0.0593$_{\pm 0.0002}$ & 0.0508$_{\pm 0.0195}$ & 0.0297$_{\pm 0.0006}$ & 0.0593$_{\pm 0.0043}$ & 0.0720$_{\pm 0.0174}$ & \underline{0.0663}$_{\pm 0.0302}$ & \textbf{0.1017}$_{\pm 0.0041}$ \\
CDs \& Vinyl & 0.0233 & 0.0129 & 0.0078 & 0.0000 & 0.0627$_{\pm 0.0153}$ & 0.0601$_{\pm 0.0113}$ & 0.0339$_{\pm 0.0057}$ & 0.0548$_{\pm 0.0115}$ & 0.0679$_{\pm 0.0074}$ & \underline{0.0762}$_{\pm 0.0113}$ & \textbf{0.1036}$_{\pm 0.0123}$ \\
Cell Phones & 0.0055 & 0.0302 & 0.0330 & 0.0110 & 0.0362$_{\pm 0.0074}$ & 0.0306$_{\pm 0.0151}$ & 0.0306$_{\pm 0.0124}$ & 0.0362$_{\pm 0.0105}$ & 0.0864$_{\pm 0.0138}$ & \underline{0.0931}$_{\pm 0.0066}$ & \textbf{0.1105}$_{\pm 0.0088}$ \\
Digital Music & 0.0185 & 0.1343 & 0.0417 & 0.0231 & 0.1264$_{\pm 0.0102}$ & 0.1149$_{\pm 0.0009}$ & 0.1322$_{\pm 0.0124}$ & 0.1437$_{\pm 0.0100}$ & \textbf{0.2299}$_{\pm 0.0061}$ & \underline{0.2241}$_{\pm 0.0119}$ & 0.1021$_{\pm 0.0226}$ \\
Electronics & 0.0100 & \underline{0.1371} & 0.1122 & 0.0175 & 0.0973$_{\pm 0.0028}$ & 0.1272$_{\pm 0.0111}$ & 0.1297$_{\pm 0.0166}$ & 0.1546$_{\pm 0.0039}$ & 0.0050$_{\pm 0.0081}$ & 0.0050$_{\pm 0.0044}$ & \textbf{0.1398}$_{\pm 0.0100}$ \\
Fashion & 0.0140 & 0.0681 & 0.0631 & 0.0301 & 0.0481$_{\pm 0.0063}$ & 0.0441$_{\pm 0.0015}$ & 0.0827$_{\pm 0.0196}$ & 0.0459$_{\pm 0.0101}$ & 0.0979$_{\pm 0.0195}$ & \underline{0.0952}$_{\pm 0.0229}$ & \textbf{0.1421}$_{\pm 0.0215}$ \\
Grocery & 0.0034 & 0.0137 & 0.0103 & 0.0103 & 0.0000$_{\pm 0.0000}$ & 0.0070$_{\pm 0.0106}$ & 0.0070$_{\pm 0.0105}$ & 0.0139$_{\pm 0.0186}$ & 0.0801$_{\pm 0.0054}$ & \underline{0.0836}$_{\pm 0.0158}$ & \textbf{0.1370}$_{\pm 0.0092}$ \\
Handmade & 0.0167 & 0.0639 & 0.0500 & 0.0389 & 0.0732$_{\pm 0.0109}$ & 0.0557$_{\pm 0.0019}$ & 0.0592$_{\pm 0.0074}$ & 0.1359$_{\pm 0.0186}$ & 0.1295$_{\pm 0.0138}$ & \underline{0.1117}$_{\pm 0.0290}$ & \textbf{0.1686}$_{\pm 0.0234}$ \\
Health & 0.0110 & 0.0495 & 0.0549 & 0.0165 & 0.0278$_{\pm 0.0039}$ & 0.0389$_{\pm 0.0125}$ & 0.0444$_{\pm 0.0167}$ & 0.0434$_{\pm 0.0007}$ & 0.0907$_{\pm 0.0187}$ & \underline{0.0980}$_{\pm 0.0103}$ & \textbf{0.1222}$_{\pm 0.0118}$ \\
Home & 0.0479 & 0.2166 & 0.1814 & 0.0957 & 0.1184$_{\pm 0.0068}$ & 0.0932$_{\pm 0.0109}$ & 0.2030$_{\pm 0.0075}$ & 0.1089$_{\pm 0.0071}$ & \underline{0.3224}$_{\pm 0.0023}$ & \textbf{0.3325}$_{\pm 0.0054}$ & 0.1371$_{\pm 0.0103}$ \\
Industrial & 0.0099 & 0.0471 & 0.0521 & 0.0074 & 0.0627$_{\pm 0.0173}$ & 0.0526$_{\pm 0.0147}$ & 0.0351$_{\pm 0.0049}$ & 0.0576$_{\pm 0.0191}$ & \underline{0.0795}$_{\pm 0.0157}$ & 0.0923$_{\pm 0.0036}$ & \textbf{0.1178}$_{\pm 0.0199}$ \\
Movies \& TV & 0.0306 & 0.0393 & 0.0087 & 0.0175 & 0.0349$_{\pm 0.0133}$ & 0.0306$_{\pm 0.0072}$ & 0.0437$_{\pm 0.0150}$ & 0.0439$_{\pm 0.0081}$ & 0.0905$_{\pm 0.0176}$ & \underline{0.0956}$_{\pm 0.0133}$ & \textbf{0.1339}$_{\pm 0.0222}$ \\
Musical Instruments & 0.0166 & 0.1140 & 0.1116 & 0.0214 & 0.1140$_{\pm 0.0083}$ & 0.1283$_{\pm 0.0171}$ & 0.1140$_{\pm 0.0157}$ & 0.1568$_{\pm 0.0042}$ & 0.0214$_{\pm 0.0197}$ & 0.0285$_{\pm 0.0158}$ & \textbf{0.1473}$_{\pm 0.0328}$ \\
Office Prod. & 0.0034 & 0.0172 & 0.0103 & 0.0172 & 0.0103$_{\pm 0.0128}$ & 0.0105$_{\pm 0.0181}$ & 0.0034$_{\pm 0.0058}$ & 0.0206$_{\pm 0.0127}$ & \underline{0.0653}$_{\pm 0.0162}$ & 0.0609$_{\pm 0.0237}$ & \textbf{0.0951}$_{\pm 0.0107}$ \\
Patio \& Garden & 0.0000 & 0.0259 & 0.0172 & 0.0086 & 0.0431$_{\pm 0.0168}$ & 0.0433$_{\pm 0.0027}$ & 0.0437$_{\pm 0.0043}$ & 0.0517$_{\pm 0.0134}$ & 0.0690$_{\pm 0.0347}$ & \underline{0.1034}$_{\pm 0.0291}$ & \textbf{0.1609}$_{\pm 0.0284}$ \\
Pet Supplies & 0.0057 & 0.0209 & 0.0152 & 0.0076 & 0.0554$_{\pm 0.0038}$ & 0.0561$_{\pm 0.0011}$ & 0.0402$_{\pm 0.0042}$ & 0.0535$_{\pm 0.0032}$ & 0.0937$_{\pm 0.0133}$ & \underline{0.0889}$_{\pm 0.0176}$ & \textbf{0.1128}$_{\pm 0.0052}$ \\
Software & 0.0288 & 0.1439 & 0.1187 & 0.0397 & 0.1572$_{\pm 0.0115}$ & 0.1703$_{\pm 0.0208}$ & 0.1965$_{\pm 0.0057}$ & 0.2227$_{\pm 0.0186}$ & \underline{0.2358}$_{\pm 0.0020}$ & \textbf{0.2882}$_{\pm 0.0063}$ & 0.0946$_{\pm 0.0186}$ \\
Sports & 0.0000 & 0.0280 & 0.0312 & 0.0218 & 0.0187$_{\pm 0.0039}$ & 0.0156$_{\pm 0.0196}$ & 0.0187$_{\pm 0.0097}$ & 0.0218$_{\pm 0.0140}$ & 0.0561$_{\pm 0.0145}$ & \underline{0.0717}$_{\pm 0.0264}$ & \textbf{0.1194}$_{\pm 0.0186}$ \\
Toys \& Games & 0.0000 & 0.0104 & 0.0000 & 0.0052 & 0.0208$_{\pm 0.0148}$ & 0.0209$_{\pm 0.0069}$ & 0.0052$_{\pm 0.0053}$ & 0.0211$_{\pm 0.0037}$ & 0.0840$_{\pm 0.0043}$ & \underline{0.0869}$_{\pm 0.0024}$ & \textbf{0.1042}$_{\pm 0.0146}$ \\
Video Games & 0.0151 & 0.0695 & 0.0658 & 0.0110 & 0.0945$_{\pm 0.0066}$ & 0.0815$_{\pm 0.0182}$ & 0.0825$_{\pm 0.0127}$ & 0.0850$_{\pm 0.0052}$ & 0.0430$_{\pm 0.0129}$ & 0.0470$_{\pm 0.0158}$ & \textbf{0.1222}$_{\pm 0.0050}$ \\
\cmidrule(lr){1-12}
\textbf{Avg.} & 0.0142 & 0.0603 & 0.0483 & 0.0195 & 0.0633$_{\pm 0.0083}$ & 0.0629$_{\pm 0.0112}$ & 0.0683$_{\pm 0.0108}$ & 0.0753$_{\pm 0.0098}$ & 0.1025$_{\pm 0.0140}$ & \underline{0.1085}$_{\pm 0.0143}$ & \textbf{0.1236}$_{\pm 0.0165}$ \\
\bottomrule
\end{tabular}

\vspace{1em}
\textbf{Panel B: NDCG@5 and MRR.}

\begin{tabular}{l cccc c cccc c}
\toprule
& \multicolumn{5}{c}{NDCG@5 $\uparrow$} & \multicolumn{5}{c}{MRR $\uparrow$} \\
\cmidrule(lr){2-6} \cmidrule(lr){7-11}
& \multicolumn{4}{c}{Retrieval} & \multicolumn{1}{c}{Ours}
& \multicolumn{4}{c}{Retrieval} & \multicolumn{1}{c}{Ours} \\
\cmidrule(lr){2-5} \cmidrule(lr){6-6} \cmidrule(lr){7-10} \cmidrule(lr){11-11}
Domain & BM25~\cite{robertson2009probabilistic} & SBERT~\cite{reimers2019sentence} & RoB.~\cite{liu2019roberta} & SimCSE~\cite{gao2021simcse} & LightCRS
& BM25~\cite{robertson2009probabilistic} & SBERT~\cite{reimers2019sentence} & RoB.~\cite{liu2019roberta} & SimCSE~\cite{gao2021simcse} & LightCRS \\
\midrule
Appliances & 0.0145 & \underline{0.1351} & 0.0863 & 0.0053 & \textbf{0.3020}$_{\pm 0.0091}$ & 0.1799 & 0.2248 & \underline{0.2454} & 0.1720 & \textbf{0.2924}$_{\pm 0.0110}$ \\
Arts \& Crafts & 0.0093 & \underline{0.0604} & 0.0437 & 0.0328 & \textbf{0.2994}$_{\pm 0.0267}$ & 0.1681 & \underline{0.2245} & 0.2065 & 0.1969 & \textbf{0.3016}$_{\pm 0.0226}$ \\
Automotive & 0.0066 & 0.0416 & \underline{0.0514} & 0.0314 & \textbf{0.3295}$_{\pm 0.0148}$ & 0.1733 & 0.2061 & \underline{0.2159} & 0.1958 & \textbf{0.3086}$_{\pm 0.0160}$ \\
Baby Products & 0.0995 & \underline{0.1302} & 0.0978 & 0.0733 & \textbf{0.3720}$_{\pm 0.0149}$ & 0.1564 & \underline{0.1591} & 0.1430 & 0.1188 & \textbf{0.3472}$_{\pm 0.0151}$ \\
Beauty & 0.1326 & 0.1678 & \underline{0.1696} & 0.1588 & \textbf{0.3617}$_{\pm 0.0145}$ & 0.1389 & \underline{0.1632} & 0.1607 & 0.1584 & \textbf{0.3472}$_{\pm 0.0152}$ \\
Books & 0.0184 & \underline{0.0917} & 0.0553 & 0.0211 & \textbf{0.3007}$_{\pm 0.0114}$ & 0.1724 & \underline{0.2224} & 0.2099 & 0.1871 & \textbf{0.2984}$_{\pm 0.0081}$ \\
CDs \& Vinyl & 0.0359 & \underline{0.0368} & 0.0286 & 0.0026 & \textbf{0.2971}$_{\pm 0.0279}$ & 0.1987 & \underline{0.2013} & 0.1940 & 0.1693 & \textbf{0.2954}$_{\pm 0.0138}$ \\
Cell Phones & 0.0141 & \underline{0.0708} & 0.0673 & 0.0307 & \textbf{0.2939}$_{\pm 0.0075}$ & 0.1799 & \underline{0.2324} & 0.2172 & 0.1812 & \textbf{0.2980}$_{\pm 0.0076}$ \\
Digital Music & 0.0348 & \underline{0.1919} & 0.0966 & 0.0339 & \textbf{0.2993}$_{\pm 0.0202}$ & 0.1984 & \textbf{0.3362} & 0.2563 & 0.1967 & \underline{0.2981}$_{\pm 0.0197}$ \\
Electronics & 0.0225 & \underline{0.2367} & 0.2025 & 0.0415 & \textbf{0.3584}$_{\pm 0.0235}$ & 0.1090 & \underline{0.2234} & 0.2010 & 0.1158 & \textbf{0.3328}$_{\pm 0.0163}$ \\
Fashion & 0.1239 & \underline{0.1952} & 0.1774 & 0.1573 & \textbf{0.3682}$_{\pm 0.0235}$ & 0.1291 & \underline{0.1845} & 0.1745 & 0.1535 & \textbf{0.3473}$_{\pm 0.0207}$ \\
Grocery & 0.0143 & \underline{0.0342} & 0.0187 & 0.0219 & \textbf{0.3194}$_{\pm 0.0054}$ & \underline{0.1648} & 0.1307 & 0.1302 & 0.1444 & \textbf{0.3184}$_{\pm 0.0058}$ \\
Handmade & 0.0386 & 0.1176 & \underline{0.1323} & 0.0728 & \textbf{0.3044}$_{\pm 0.0207}$ & 0.2025 & 0.2736 & \underline{0.2906} & 0.2330 & \textbf{0.2969}$_{\pm 0.0157}$ \\
Health & 0.0357 & 0.1196 & \underline{0.1334} & 0.0536 & \textbf{0.3095}$_{\pm 0.0106}$ & 0.2005 & \underline{0.2371} & 0.2186 & 0.2175 & \textbf{0.3061}$_{\pm 0.0117}$ \\
Home & 0.2417 & \textbf{0.4017} & \underline{0.3864} & 0.3147 & 0.3584$_{\pm 0.0168}$ & 0.2505 & \textbf{0.3700} & \underline{0.3696} & 0.3137 & 0.3384$_{\pm 0.0094}$ \\
Industrial & 0.0320 & 0.1017 & \underline{0.1036} & 0.0319 & \textbf{0.3088}$_{\pm 0.0095}$ & 0.1831 & \underline{0.2260} & 0.2110 & 0.1949 & \textbf{0.3100}$_{\pm 0.0113}$ \\
Movies \& TV & 0.0462 & \underline{0.0745} & 0.0323 & 0.0312 & \textbf{0.3235}$_{\pm 0.0419}$ & 0.1886 & \underline{0.2285} & 0.1975 & 0.1950 & \textbf{0.3183}$_{\pm 0.0253}$ \\
Musical Instruments & 0.0220 & 0.1754 & \underline{0.1841} & 0.0401 & \textbf{0.3716}$_{\pm 0.0268}$ & 0.1149 & \underline{0.2026} & 0.2004 & 0.1193 & \textbf{0.3514}$_{\pm 0.0268}$ \\
Office Prod. & 0.0183 & \underline{0.0472} & 0.0430 & 0.0322 & \textbf{0.2955}$_{\pm 0.0050}$ & 0.1824 & \underline{0.2085} & 0.2006 & 0.1960 & \textbf{0.2948}$_{\pm 0.0087}$ \\
Patio \& Garden & 0.0100 & 0.0647 & \underline{0.0648} & 0.0317 & \textbf{0.3249}$_{\pm 0.0177}$ & 0.1767 & 0.2271 & \underline{0.2286} & 0.1969 & \textbf{0.3305}$_{\pm 0.0141}$ \\
Pet Supplies & 0.0177 & \underline{0.0668} & 0.0570 & 0.0279 & \textbf{0.2953}$_{\pm 0.0069}$ & 0.1758 & \underline{0.2300} & 0.2159 & 0.1933 & \textbf{0.3000}$_{\pm 0.0028}$ \\
Software & 0.0474 & \underline{0.2670} & 0.2358 & 0.0966 & \textbf{0.2693}$_{\pm 0.0073}$ & 0.2093 & \textbf{0.3993} & \underline{0.3574} & 0.2567 & 0.2738$_{\pm 0.0124}$ \\
Sports & 0.0142 & \underline{0.0651} & 0.0565 & 0.0330 & \textbf{0.3130}$_{\pm 0.0199}$ & 0.1809 & \underline{0.2188} & 0.2180 & 0.1960 & \textbf{0.3116}$_{\pm 0.0153}$ \\
Toys \& Games & 0.0104 & \underline{0.0462} & 0.0225 & 0.0212 & \textbf{0.3087}$_{\pm 0.0032}$ & 0.1629 & \underline{0.2111} & 0.1892 & 0.1794 & \textbf{0.2986}$_{\pm 0.0046}$ \\
Video Games & 0.0234 & \underline{0.1210} & 0.1149 & 0.1099 & \textbf{0.3451}$_{\pm 0.0060}$ & 0.1136 & \underline{0.1626} & 0.1592 & 0.1099 & \textbf{0.3248}$_{\pm 0.0021}$ \\
\cmidrule(lr){1-11}
\textbf{Avg.} & 0.0434 & \underline{0.1224} & 0.1065 & 0.0567 & \textbf{0.3212}$_{\pm 0.0157}$ & 0.1724 & \underline{0.2282} & 0.2164 & 0.1837 & \textbf{0.3136}$_{\pm 0.0133}$ \\
\bottomrule
\end{tabular}
\end{table*}

\section{Prompt Templates}
\label{sec:appendix-prompts}

This appendix consolidates the four prompt templates that govern our data curation and model training pipeline. Table~\ref{tab:prompt-filter} isolates Reddit threads that end in an explicit product recommendation, Table~\ref{tab:prompt-caption} elicits visually grounded captions, and Tables~\ref{tab:prompt-rec-text}--\ref{tab:prompt-rec-image} drive the recommendation step in the text-only and the text-plus-image configurations, respectively.




\begin{table*}[t]
\centering
\caption{Prompt used to identify product-recommendation conversations. Only conversations labeled \texttt{yes} by GPT-3.5-turbo are retained in \textit{MM-R2A-25}.}
\label{tab:prompt-filter}
\begin{tabular}{p{0.95\linewidth}}
\toprule
You are given a Reddit conversation containing multiple numbered messages. Each message follows the format
\texttt{\textless message\textgreater(i) [content]\textless/message\textgreater}.

Determine whether the final message explicitly recommends one or more specific items in response to a recommendation request in an earlier message.

Answer \texttt{yes} if both conditions are satisfied, \texttt{no} otherwise, or \texttt{uncertain} if the conversation does not provide enough information.

Return only \texttt{yes}, \texttt{no}, or \texttt{uncertain}, without any explanation.

\textbf{Conversation:} \texttt{\{conversation\}}
\\
\bottomrule

\end{tabular}
\end{table*}

\begin{table*}[t]
\centering
\caption{Prompt used to generate visually grounded product descriptions.}
\label{tab:prompt-caption}
\begin{tabular}{p{0.95\linewidth}}
\toprule
You are given an Amazon product title and image. Generate a detailed, visually grounded description that helps users evaluate the product. Focus on its appearance, visible features, and other visually informative details. Do not repeat or mention the product title.

\textbf{Product title:} \texttt{\{title\}}

\textbf{Product image:} \texttt{\{image\}}
\\
\bottomrule

\end{tabular}
\end{table*}

\begin{table*}[t]
\centering
\caption{Prompt for text-only product recommendation, used by generative baselines.}
\label{tab:prompt-rec-text}
\begin{tabular}{p{0.95\linewidth}}
\toprule
You are given a conversation between a user requesting product recommendations, marked by \texttt{\textless submission\textgreater}, and other users providing responses, marked by \texttt{\textless comment\textgreater}. You are also given candidate products formatted as \texttt{ID: title}.

Select the candidate that best matches the user's needs. Return only its ID without any explanation.

\textbf{Conversation:} \texttt{\{conversation\}}

\textbf{Candidates:} \texttt{\{list of candidates\}}
\\
\bottomrule

\end{tabular}
\end{table*}

\begin{table*}[t]
\centering
\caption{Prompt for image-aware product recommendation, used by generative baselines.}
\label{tab:prompt-rec-image}
\begin{tabular}{p{0.95\linewidth}}
\toprule
You are given a conversation between a user requesting product recommendations, marked by \texttt{\textless submission\textgreater}, and other users providing responses, marked by \texttt{\textless comment\textgreater}. You are also given candidate products, each formatted as \texttt{ID: title} followed by its image.

Select the candidate that best matches the user's needs based on the conversation, title, and image. Return only its ID without any explanation.

\textbf{Conversation:} \texttt{\{conversation\}}

\textbf{Candidates:} \texttt{\{list of candidates\}}
\\
\bottomrule

\end{tabular}
\end{table*}

\section{Hyperparameter Details}
\label{sec:appendix-hyperparameters}

No domain-specific hyperparameter tuning was conducted. We used one fixed configuration across all 25 product domains and trained the model (Section~\ref{sec:method}) under this unified protocol.

The common projection dimension is $D = 256$. The query--candidate cross-attention stack has $L = 2$ layers with 4 attention heads and a feed-forward dimension of 1024 ($4 \times D$), and the inter-candidate comparison Transformer mirrors this configuration (2 layers, 4 heads, 1024 FFN). The scoring head is a 2-layer MLP ($D \to D/2 \to 1$), and dropout is fixed at 0.1 across all trainable modules. Frozen encoder outputs feed in at their native hidden sizes: $d_q = 768$ for the GTE-base text encoder and $d_v = 4096$ for the LLaVA-v1.6 visual tokens, with a maximum of $N = 15$ image descriptions and 5 visual tokens per candidate item. The training objective is a cross-entropy classification loss over the $K$-way candidate set. Models are trained for up to 20 epochs with early stopping (patience $= 2$) on validation loss, using a batch size of 16 and the AdamW optimizer ($\text{lr} = 1 \times 10^{-4}$, weight decay $1 \times 10^{-5}$) under a constant learning-rate schedule with no warmup or decay. Gradient clipping (max norm 1.0) and BF16 mixed precision are applied throughout.

\section{Offline Preprocessing Cost and Storage Footprint}
\label{sec:appendix-offline-cost}

LightCRS moves product-side multimodal encoding from online scoring to offline preprocessing (Section~\ref{sec:offline-features}). This appendix reports the computational cost and storage overhead of this offline process.

\paragraph{Measurement methodology.}
All measurements are conducted on a single NVIDIA DGX Spark workstation. We evaluate four offline processing steps: text encoding(title/features/descriptions), image caption generation, caption encoding, and visual-token extraction. Each step is measured independently. We sample 50 products from the Books category, measure the runtime and peak GPU memory after one warm-up pass, and linearly extrapolate the results to the full MM-R2A-25 catalog since each product is processed independently.

\paragraph{Offline preprocessing cost.}
Table~\ref{tab:offline-cost} reports the runtime, throughput, peak GPU memory, and extrapolated cost on the full catalog. Caption generation dominates the preprocessing cost, requiring 593.46 GPU-hours for generating captions of all product images, while the remaining encoding stages require less than 3 GPU-hours in total. This cost is incurred only once per catalog. After preprocessing, the online module operates only on cached representations without additional product-side encoding.

\paragraph{Storage footprint.}
Table~\ref{tab:cache-storage} summarizes the storage cost of cached product representations. We notice that visual tokens accounts for the majority of storage usage. 

\begin{table*}[t]
\centering
\caption{Per-stage offline preprocessing cost measurement. ``Wall-clock'' denotes the measured runtime on the 50-item sample after one warm-up pass; ``Throughput'' denotes the processing rate in items or encodings per second; ``Peak GPU mem'' denotes the maximum allocated GPU memory during the timed run; ``Full MM-R2A-25'' reports the extrapolated cost for 50{,}121 products and 219{,}268 product images.}
\label{tab:offline-cost}
\setlength{\tabcolsep}{6pt}
\renewcommand{\arraystretch}{1.15}
\resizebox{\linewidth}{!}{%
\begin{tabular}{@{}rlllrrrr@{}}
\toprule
\textbf{\#} & \textbf{Stage} & \textbf{View} & \textbf{Encoder} & \textbf{Wall-clock} & \textbf{Throughput} & \textbf{Peak GPU mem} & \textbf{Full MM-R2A-25} \\
 & & & & (50 items) & (items/s) & (GB) & (GPU-hours) \\
\midrule
1 & Text encoding          & title / features / description & GTE-base                    & 0.47~s   & 319.3 & 0.29  & 0.13    \\
2 & Caption generation     & image captions                 & LLaVA-v1.6 (full VLM)       & 487.18~s & 0.10  & 15.03 & 593.46  \\
3 & Caption encoding       & caption embeddings             & GTE-base                    & 0.18~s   & 277.3 & 0.28  & 0.22    \\
4 & Visual extraction      & visual tokens                  & LLaVA-v1.6 vision tower     & 9.32~s   & 5.36  & 14.43 & 2.60    \\
\bottomrule
\end{tabular}%
}
\end{table*}

\begin{table*}[t]
\centering
\caption{Cached product-representation footprint. Each view is stored as a tensor with the indicated shape and FP32 precision. Per-product KB uses the catalog-wide average of 4.38~images per product for the caption row; full-catalog totals apply this same average to all products.}
\label{tab:cache-storage}
\setlength{\tabcolsep}{8pt}
\renewcommand{\arraystretch}{1.15}
\begin{tabular}{@{}lllrr@{}}
\toprule
\textbf{View}                & \textbf{Tensor shape} & \textbf{Source}                       & \textbf{Per-product} & \textbf{Full MM-R2A-25} \\
                             &                       &                                       & \textbf{(KB, FP32)}  & \textbf{(GB, FP32)}     \\
\midrule
Title embedding              & $(768,)$              & GTE-base                              & 3.1   & 0.15  \\
Features embedding           & $(768,)$              & GTE-base                              & 3.1   & 0.15  \\
Description embedding        & $(768,)$              & GTE-base                              & 3.1   & 0.15  \\
Caption embeddings           & $(N, 768)$            & GTE-base over LLaVA-v1.6 captions     & 13.5  & 0.68  \\
Visual tokens                & $(5, 4096)$           & LLaVA-v1.6 vision tower               & 81.9  & 4.11  \\
\bottomrule
\end{tabular}
\end{table*}

\end{document}